\documentclass[utf8]{FrontiersinHarvard} % for articles in journals using the Harvard Referencing Style (Author-Date), for Frontiers Reference Styles by Journal: https://zendesk.frontiersin.org/hc/en-us/articles/360017860337-Frontiers-Reference-Styles-by-Journal
\usepackage{url,hyperref,lineno,microtype,subcaption}
\usepackage[onehalfspacing]{setspace}
\usepackage{csquotes}
\usepackage{amsmath}
\usepackage{amsfonts}
\usepackage{caption}
\usepackage{hyperref}
\usepackage{booktabs} % horizontal lines in tables
\usepackage{./preamble}
\usepackage[T1]{fontenc}
\usepackage{placeins}

\theoremstyle{definition}

\newenvironment{deftable}[1]
{
    \begin{minipage}{\textwidth}
        \begin{tabular}{cp{3.25in}}
            \multicolumn{2}{l}{\textbf{#1}} \\
            }
            {
        \end{tabular}
    \end{minipage}
}

\def\keyFont{\fontsize{8}{11}\helveticabold }
\def\firstAuthorLast{Muhr {et~al.}} %use et al only if is more than 1 author
\def\Authors{David Muhr\,$^{1,2}$, Michael Affenzeller\,$^{2,3}$ and Josef Küng\,$^{4}$}
\def\Address{
$^{1}$BMW Group, Steyr, Austria,
$^{2}$Institute for Formal Models and Verification, Johannes Kepler University, Linz, Austria
$^{3}$ Heuristic and Evolutionary Algorithms Laboratory, University of Applied Sciences Upper Austria, Hagenberg, Austria
$^{4}$Institute for Application-oriented Knowledge Processing, Johannes Kepler University, Linz, Austria}
\def\corrAuthor{David Muhr}
\def\corrEmail{david.muhr@bmw.com}

\begin{document}
\onecolumn
\firstpage{1}

\title {A Probabilistic Transformation of Distance-Based Outliers}

\author[\firstAuthorLast ]{\Authors} %This field will be automatically populated
\address{} %This field will be automatically populated
\correspondance{} %This field will be automatically populated
\extraAuth{}% If there are more than 1 corresponding author, comment this line and uncomment the next one.
%\extraAuth{corresponding Author2 \\ Laboratory X2, Institute X2, Department X2, Organization X2, Street X2, City X2 , State XX2 (only USA, Canada and Australia), Zip Code2, X2 Country X2, email2@uni2.edu}

\maketitle

\begin{abstract}
	% For full guidelines regarding your manuscript please refer to \href{https://www.frontiersin.org/guidelines/author-guidelines}{Author Guidelines}.
	% As a primary goal, the abstract should render the general significance and conceptual advance of the work clearly accessible to a broad readership. References should not be cited in the abstract. Leave the Abstract empty if your article does not require one, please see the Article Types on every Frontiers journal page for full details

	% What have the authors done (essence) ? 1-2 sentences.
	The scores of distance-based outlier detection methods are difficult to
	interpret, making it challenging to determine a cut-off threshold
	between normal and outlier data points without additional context.
	We describe a generic transformation of distance-based outlier scores into
	interpretable, probabilistic estimates.
	The transformation is ranking-stable and increases the contrast between
	normal and outlier data points.
	% Why have they done it? If necessary or you have space. 1-2 sentences.
	% How have they done it? 3-4 sentences
	Determining distance relationships between data points is necessary to
	identify the nearest-neighbor relationships in the data, yet, most of the
	computed distances are typically discarded.
	We show that the distances to other data points can be used to model
	distance probability distributions and, subsequently, use the distributions
	to turn distance-based outlier scores into outlier probabilities.
	% What are the main results? With numbers if available, 3-4 sentences
	Our experiments show that the probabilistic transformation does not impact
	detection performance over numerous tabular and image benchmark datasets but
	results in interpretable outlier scores with increased contrast between
	normal and outlier samples.
	% What is the importance and impact of the results?, 1-2 sentences
	Our work generalizes to a wide range of distance-based outlier detection
	methods, and because existing distance computations are used, it adds no
	significant computational overhead.

	\tiny
	\keyFont{ \section{Keywords:} anomaly detection, outlier detection, outlier score, anomaly score, score normalization}
\end{abstract}

\section{Introduction}

% For Original Research Articles \citep{conference}, Clinical Trial Articles \citep{article}, and Technology Reports \citep{patent}, the introduction should be succinct, with no subheadings \citep{book}. For Case Reports the Introduction should include symptoms at presentation \citep{chapter}, physical exams and lab results \citep{dataset}.

% What has been done? (brief) Immediate purpose statement.

We propose a generic method to transform distance-based outlier detection
models into interpretable, probabilistic models.
An outlier is often described as \enquote{an observation (or subset of
	observations) which appears to be inconsistent with the remainder of that set
	of data} \cite{barnettOutliersStatisticalData1978}.
The definition of an \enquote{inconsistent} observation is not uniform and
varies depending on the application and algorithm used.
Inconsistency can mean that the outlier object stems from a different
distribution than the model describing the data, which reflects the
classical definition of outliers by \cite{hawkinsIdentificationOutliers1980}:
\enquote{An outlier is an observation which deviates so much from the other
	observations as to arouse suspicions that it was generated by a different
	mechanism.
}.
An outlier is also referred to as an anomaly or novelty, sometimes
interchangeably.
Therefore, outlier detection is also referred to as anomaly detection or
novelty detection.
Because the methods used to detect outliers, anomalies, and novelties are
mostly the same, we make no distinction between these terms and refer to
inconsistent instances as outliers.
In a distance-based setting, we can define outliers as objects located
far away from the remaining objects.

\begin{deftable}{Notation}
    $\displaystyle a$		      & A scalar (integer or real)
    \\
    $\displaystyle \va$ 	      & A vector
    \\
    $\displaystyle \mA$ 	      & A matrix
    \\
    $\displaystyle \ra$ 	      & A scalar random variable
    \\
	$\displaystyle \sA$ 		  & A set
	\\
	$\displaystyle \spA$ 	  	  & A space
	\\
	$\displaystyle A$			  & A distribution
	\\
    $\displaystyle \R$			  & The set of real numbers
	\\
	$\displaystyle \sX$ 	  	  & A dataset
	\\
	$\displaystyle \{0, 1, \dots, n \}$ & The set of all integers between $0$
    and $n$
	\\
	$\displaystyle f(x): \sA \rightarrow \sB$ & A function of $x$ with domain $\sA$
    and range $\sB$
\end{deftable}

Specifically, given a metric space $(\spM, d)$ with metric $d$, each
object $\vx \in \spM$ receives a real-valued outlier score $s := q(\vx)$ via a
function $q: \spM \to \sR$, where the function depends on the distances to the
other objects in the dataset.
To determine if an observation is considered an outlier, it is necessary to
to establish a threshold value converting outlier scores into binary labels of
normal and outlier data points.
% Why has it been done? Motivation and scope of the paper
A major challenge in distance-based outlier detection is the interpretation of
the resulting scores.
The scores provided by distance-based methods differ widely in their scale,
range, and meaning.
Even when considering only a single outlier detection method, the same outlier
score can describe different degrees of outlierness depending on the kind of
data.
These challenges make the interpretation and comparison of outlier scores
difficult.
% What is the state of the art? Critical literature review, not just listing.
% Review of outlier detection normalization.
% What are the objectives? (detailed) Paper’s objectives.
Distance-based outlier detection scores are typically derived from some
neighborhood representation given a distance matrix.
We propose that the information contained in the distance matrix can be used to
derive a probabilistic normalization of outlier scores such that they become
interpretable.
% Show closed-world to open-world.
% How do you test the hypothesis? Overall methodology.
Based on a large number of benchmark datasets, we test our approach in terms of
detection performance and interpretability and show that it is possible to
achieve interpretable, probabilistic outlier scores with no detriment to the
resulting detection performance.
% TODO: add references?% Did your hypothesis succeed? Brief evaluation.
% How is the paper structured? Paper outline.
The rest of this paper is organized as follows.
Section \ref{sec:distance} provides an overview of distance-based outlier
detection methods.
In Section \ref{sec:normalization}, we show score normalization schemes and
their application to distance-based methods.
In Section \ref{sec:probabilistic}, we describe our proposed probabilistic
normalization scheme, and in Section \ref{sec:results}, we describe the results
of applying our scheme on benchmark datasets.
Finally, in Section \ref{sec:conclusion}, we derive conclusions and provide
opportunities for future research.

\section{Distance-based Outlier Detection} \label{sec:distance}

In this section, we introduce and review common distance-based outlier detection
methods and formalize them as a scoring function $q : \spM \to \sR$ on a metric
space $(\spM, d)$, such that an outlier detection method assigns a real-valued
outlier score to an observation.
We further differentiate between the closed-world and open-world outlier
detection setting, an often disregarded yet highly relevant aspect of
distance-based outlier detection.
The following outlier detection methods are formulated in a closed-world setting,
such that the observations in a dataset $\train$ are assigned an outlier score.
Often, however, it is necessary to assign an outlier score to unseen data, such
that a model of normality is determined based on a dataset $\train$,
and the outlier score is determined on unseen observations in a dataset $\test$.
At the end of this section, we provide a simple approach to transfer said
closed-world outlier detection methods into an open-world setting.

\subsection{$k$-th Nearest Neighbors}

\cite{knorrUnifiedApproachMining1997, knorrAlgorithmsMiningDistanceBased1998, knorrDistancebasedOutliersAlgorithms2000}
first formalized a distanced-based notion of outliers in which an object
$\vx \in \train$ is said to be a DB-outlier in a dataset of $n$ objects
if $\left|\{ \vx' \in \train \mid d(\vx, \vx') > \delta \}\right| \geq \alpha n$
where $\alpha, \delta \in \sR$ are parameters to be specified by the user and
$0 \leq \alpha \leq 1$.
In this specification, a fraction $\alpha$ of all objects have a distance from
$\vx$ that is larger than $\delta$.
\cite{chandolaAnomalyDetection2009} point out that this method can be viewed as
global density estimation for each instance since it involves counting
the number of neighbors in a hypersphere of radius $\delta$.
However, a major drawback of this definition is that it is difficult to
determine a distance threshold $\delta$ and that the results do not determine a
ranking of scores.

\cite{ramaswamyEfficientAlgorithmsMining2000} build upon the ideas presented in
DB-outliers.
To determine the outlier score of an instance, they propose to use the distance
to its $\kth$-nearest neighbor as a score; thus, we refer to the method as
$\kthNN$.
Compared to DB-outliers, the main benefit of this approach is that it does not
require the user to specify a distance $\delta$.
The $\kthNN$ outlier score of an observation $\vx$ is defined as

\begin{equation}
	q_{\kthNN}(\vx) := d^{(k)}(\vx, \train)
\end{equation}

where $\vx \in \train$ and $d^k(\vx,
	\train)$ is the distance between $\vx$ and its $\kth$ nearest neighbor in
$\train$.

\cite{angiulliFastOutlierDetection2002} adapt the $\kthNN$ approach to use
the average distance to the $k$-nearest neighbors of a point $\vx$ instead of
the $\kth$ distance, which can also be interpreted as the maximum distance.
We refer to this method as $\kNN$ and define it as follows

\begin{equation}
	q_{\kNN}(\vx) := \frac{1}{k} \sum_{i = 1}^k d^{(i)}(\vx,
	\train)
\end{equation}

where $\vx \in \train$ and
$d^{(i)}(\vx, \train)$ is the distance between $\vx$ and its $i^{\mathrm{th}}$
nearest neighbor in $\train$.

\begin{figure}[h!]
	\centering
	\includegraphics[width=\linewidth]{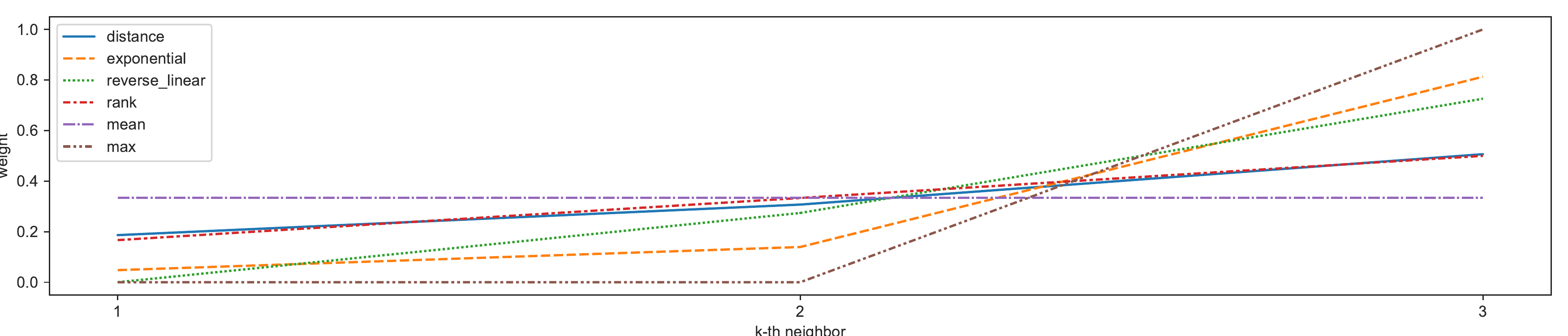}
	\caption{ Different weighting schemes for three nearest neighbors with
		distances $\exp([0.5, 1.0, 1.5])$ and fixed hyperparameters
		$s = a = b = 1$. }
	\label{fig:weighting}
\end{figure}

We propose to generalize $\kthNN$ and $\kNN$ as specific instances of weighting
schemes for distance-based outlier detection.
Weighting schemes are commonly used in $k$-nearest neighbors classification,
where the schemes traditionally emphasize close neighbors and disregard
neighbors farther away \cite{gelerComparisonDifferentWeighting2016}.
However, as evident in $\kthNN$-based outlier detection, where only the
farthest neighbor is considered, we propose to emphasize the neighbors
farther away.
A further difference between weighted-neighbors classification and outlier
is the predicted result, which corresponds to class votes or outlier scores.
To keep the resulting outlier scores in the same range, we propose to
sum-normalize the weights such that the resulting weight vectors sum to one.
The resulting outlier scores can subsequently be interpreted as a
(smoothened) distance or distance probability.
We adapt three of the weighting measures investigated in
\cite{gelerComparisonDifferentWeighting2016} to the outlier detection task
and describe $\kthNN$ as \textit{max}-weighted and $\kNN$ as
\textit{mean}-weighted outlier detection.
The \textit{distance} and \textit{rank} schemes are adapted from Dudani's
weighted nearest neighbor classification
\cite{dudaniDistanceWeightedKNearestNeighborRule1976}, the \textit{exponential}
scheme from \cite{zavrelEmpiricalReExaminationWeighted1997}, and the
\textit{linear} scheme from
\cite{macleodReExaminationDistanceWeightedKNearest1987}.
In all cases, we reverse the schemes such that the farthest neighbor gets the
largest weight.
We define the schemes for a vector of $k$-nearest neighbor distances $\vd$ as
follows

\begin{align}
	\vw_{\max}(\vd)                       & = [0, 0, \ldots, 1]                               					\\
	\vw_{\mathrm{mean}}(\vd)              & = \left[\frac{1}{k}, \frac{1}{k}, \ldots, \frac{1}{k}\right] 		\\
	\vw_{\mathrm{distance}}(\vd, s)       & = [d_1^s, d_2^s, \ldots, d_n^s]                                     \\
	\vw_{\mathrm{exponential}}(\vd, a, b) & = [\exp(a d_1^b), \exp(a d_2^b), \ldots, \exp(a d_n^b)]				\\
	\vw_{\mathrm{linear}}(\vd)            & = [\mathrm{norm}(d_1), \mathrm{norm}(d_2), \ldots, \mathrm{norm}(d_n)] 	  					\\
	\vw_{\mathrm{rank}}(\vd)              & = [1,2,\ldots,k]
\end{align}

where $\vw_{\max}$ is $1$ only if $d_i = \max(\vd)$,

\begin{equation} \label{eq:norm}
	\mathrm{norm} = \frac{d_i - \min(\vd)}{\max(\vd) - \min(\vd)},
\end{equation}

and $s$, $a$ and $b$ are hyperparameters of the respective weighting schemes.
We show how the weights influence the determination of an outlier score based on
a three-nearest-neighbors example in Figure \ref{fig:weighting}.

Because outlier scores are assumed to be positive values derived from distances, 
sum-normalization is possible by dividing each element in the weight vector
by its sum as defined in Equation \ref{eq:knnw}.
Sum-normalization ensures that the weight vector sums to one and the weighted
outlier score can be interpreted as a weighted distance.
We further use the proposed weighting scheme to define a generic weighted
$k$-nearest neighbor approach as $\kNNW$, which serves as a basis for our
tabular outlier detection experiments in Section \ref{sec:results}.

\begin{equation} \label{eq:knnw}
	q_{\kNNW}(\vx) := \frac{\vd \cdot \vw}{\sum_{i=1}^k w_i}
\end{equation}

where $\vd$ is the vector of $k$-nearest neighbor distances, $\vw$ is
the $k$-dimensional weight vector and $\cdot$ denotes the dot product
between the distance- and weight vector.

More recently, authors proposed various sampling schemes to improve the
efficiency of the described techniques.
\cite{wuOutlierDetectionSampling2006} propose an iterative sampling scheme to
approximate the $\kthNN$ score, which we designate as $\kth$ iteratively sampled
nearest neighbor $\kthISNN$.

\begin{equation}
	q_{\kthISNN}(\vx) := d^{(k)}(\vx, S_{\vx}(\train))
\end{equation}

where $S_{\vx}(\train)$ is a randomly sampled subset of $\train$ excluding
$\vx$.
The subsampling is determined individually for each point $\vx'$ processed with
$q_{\kthISNN}(\vx')$; therefore, it is referred to as iterative sampling.

\cite{sugiyamaRapidDistanceBasedOutlier2013} show that a simplification of
$\kthISNN$ leads to better detection performance over 16 different datasets.
The authors propose to remove the iterative aspect of $\kthISNN$ and, instead,
sample only once for all data points and identify the first nearest neighbor,
which we describe as the sampled nearest neighbor or $\SNN$.

\begin{equation}
	q_{\SNN}(\vx) := \underset{\vx' \in S(\train)}{\min} d(\vx, \vx')
\end{equation}

where $S(\train)$ is an independent random subset of the data that is
determined once.
In other words, for a point $\vx$, this method uses the distance to its closest
point $\vx'$ in a fixed sample $S(\train)$ as an outlier score.

\cite{pangLeSiNNDetectingAnomalies2015} extend the $\SNN$ approach by
repeatedly sampling random subsets of the data, which we term repeatedly
sampled nearest neighbor $\RSNN$.

\begin{equation}
	q_{\RSNN}(\vx) := \frac{1}{r} \sum_{i = 1}^{r} \underset{\vx' \in S_i(\train)}{\min} d(\vx, \vx')
\end{equation}

where $r$ is the number of random subsets to sample and $S_i(\train)$ is the
$i$-th random sample.
This method essentially represents an ensemble of nearest neighbor outlier
detection models and, therefore, expectedly improves upon $\SNN$, which the
authors empirically show using 11 datasets.
It can be argued that $k$-nearest neighbors ensembles with data subsampling are
a generalization of $\RSNN$, which are well-known techniques to improve
neighbor-based outlier detection
\cite{zimekSubsamplingEfficientEffective2013,aggarwalTheoreticalFoundationsAlgorithms2015,muhrLittleDataOften2022}.

In addition to data sampling techniques, other authors use randomized sampling
to determine feature subspaces as initially motivated by
\cite{aggarwalOutlierDetectionHigh2002}.
\cite{kriegelOutlierDetectionAxisParallel2009} define a set of reference points
based on the concept of shared nearest neighbors.
The reference points characterize a subspace hyperplane, and the outlier scores
are determined by the Euclidean distance of a point $\vx$ to the subspace
hyperplane, weighted by an indicator function that determines the relevance of
a dimension.
\cite{agrawalLocalSubspaceBased2009} proposes a very similar distance-based
subspacing approach.
\cite{zhangAnglebasedSubspaceAnomaly2015} also use a shared nearest neighbor
reference set to determine subspaces, using an angle-based approach to compute
the outlier scores.
\cite{trittenbachDimensionbasedSubspaceSearch2019} propose a method to determine
subspaces that considers the relationship between subspaces.
\cite{kellerHiCSHighContrast2012} propose to determine high-contrast subspaces
for outlier detection as a form of data pre-processing.
\cite{caberoArchetypeAnalysisNew2021} also determine the subspaces as a data
pre-processing step based on archetypal analysis followed by a $\kthNN$
approach.

Some authors combine distance-based outlier detection with dimensionality
reduction techniques such as principal component analysis
\cite{dangDistancebasedKnearestNeighbors2015} for high-dimensional data.
In image-based outlier detection, authors use neural networks to evaluate the
neighborhood search in latent spaces describing entire images
\cite{bergmanDeepNearestNeighbor2020}, image sub-features
\cite{cohenSubImageAnomalyDetection2021}, or image patch-features
\cite{rothTotalRecallIndustrial2022}.

Another option to model distance-based outliers is to use reverse nearest
neighbors or natural neighbors relationships.
For example, Outlier Detection using Indegree Number (ODIN)
\cite{hautamakiOutlierDetectionUsing2004} models the nearest neighbor
relationships as a directed graph and defines the outlier score as the in-degree
number in the graph such that a low in-degree number defines an outlier.
\cite{radovanovicReverseNearestNeighbors2015} analyze the concept of hubness,
which appears in reverse nearest neighbors outlier detection, and propose an
outlier detection method based on anti-hubs; points that do not occur in
the nearest neighbors of any other points.
Natural neighbors approaches discard the $k$-nearest neighbors parameter and
instead perform a search over $\lambda$ rounds to identify an appropriate number
of neighbors such that a shared neighbor relationship is found
\cite{zhuNaturalNeighborSelfadaptive2016,wahidNaNODNaturalNeighbourbased2021}.
A further extension is described by extended nearest neighbor approaches, which
combine the nearest neighbors with reverse nearest neighbors and shared nearest
neighbors \cite{tangENNExtendedNearest2015,tangLocalDensitybasedApproach2017}.

\subsection{Local Outlier Factor}
\label{sec:lof}

In contrast to the previously described techniques, referred to as
global outlier detection techniques, the Local Outlier Factor (LOF)
\cite{breunigLOFIdentifyingDensityBased2000} model introduces the concept of
local outliers.
\cite{schubertLocalOutlierDetection2014} formalize distance-based outlier
detection models such that an outlier score is determined based on some
\textit{context set}, typically the $k$-nearest neighbors of a point $\vx$.
To compare the resulting outlier scores, another set of points is used, which is
referred to as the \textit{reference set}.
Global methods compare the resulting score from the context set to all other
points in the reference set, the dataset $\train$.
Because the comparison of scores is global, those methods ignore differences in
the local densities of the data.
Local methods use a different reference set to compare the scores to, typically,
the $k$-nearest neighbors as in the context set.
Local methods convert the distance information from the local neighborhood into
some form of density; therefore, the methods are sometimes also referred to as
density-based.
LOF can be defined as a scoring function

\begin{equation}
	q_{\LOF}(\vx) := p(\vx)^{-1} \left(\left|N^k(\vx)\right|^{-1} \sum_{\vx' \in
		N^k(\vx)} p(\vx')\right)
\end{equation}

where $N^k(\vx)$ is
the set of $k$-nearest neighbors of $\vx$ and $p$ is the local reachability
density of $\vx$ defined as $p(\vx) := |N^k(\vx)|^{-1} \sum_{\vx' \in N^k(\vx)}
	\max \{d^k(\vx', \train), d(\vx, \vx')\}$.
Note that $N^k(\vx)$ includes all objects inside the $\kth$-distance, which can,
in the case of a \enquote{tie}, be more than $k$ objects.
The local reachability density can be seen as an average inverse distance of a
point $\vx$ normalized such that the distance cannot be smaller than
the $\kth$-distance.
According to the authors, the local reachability density stabilizes and prevents
statistical fluctuations, a fact later analyzed in more detail by
\cite{schubertGeneralizedOutlierDetection2014}.
The local outlier factor then compares $p(\vx)$, the density of the context set,
to the average reachability density of the points in the reference set.
If the average reachability density in the reference set is higher than the
point density obtained from the context set, then the score of the local
outlier factor is above one and considered less normal.

% https://github.com/elki-project/elki/blob/bcea3cbf286698867f64f4cb794268f6eb809139/elki-outlier/src/main/java/elki/outlier/lof/SimplifiedLOF.java
\cite{schubertLocalOutlierDetection2014} propose a simplified version of the
local outlier factor where $p$ is the inverse $\kth$-distance
$p(\vx) := d^k(\vx, \train)^{-1}$, which represents a simpler density
estimate as compared to the local reachability density in $\LOF$.
To better illustrate the general concept of local outlier detection, the
simplified local outlier factor ($\SLOF$) can be stated as follows

\begin{equation}
	q_{\SLOF}(\vx) := \frac{\underset{\vx' \in
		N^k(\vx)}{\mathrm{mean}}(d^k(\vx', \train)^{-1})} {d^k(\vx, \train)^{-1}}.
\end{equation}

The authors show that many local outlier models can be considered variations
of the simplified local outlier factor model.
For example, LDOF proposed by \cite{zhangNewLocalDistanceBased2009} is a
variation of the simplified LOF model using an average distance as in $\kNN$
instead of the $\kth$-distance.
Influence Outlierness (INFLO) \cite{jinRankingOutliersUsing2006} is another
variation of the simplified LOF, which diverges by using a different context
set that includes reverse nearest neighbors.
Another method that can be seen as an extension of the simplified LOF is Local
Outlier Probabilities (LoOP) \cite{kriegelLoOPLocalOutlier2009}, which adds a
probabilistic normalization to $\SLOF$.
Many more local outlier detection methods have been described in the literature
covering entire literature reviews
\cite{alghushairyReviewLocalOutlier2021}.
\cite{schubertGeneralizedOutlierDetection2014} note that local outlier detection
methods can be differentiated using their order of locality, and
\cite{goodgeLUNARUnifyingLocal2022} show that the methods can be generalized
as message-passing algorithms on a nearest neighbors graph.

\subsection{Closed-world and Open-world}

Distance-based outlier detection methods are typically defined in a closed-world
setting; however, there is an important difference between the closed-world and
open-world specification such that, for
two equal points $\vx \in \train$, $\vx' \in \test$ and $\vx = \vx'$, the $\kth$
nearest neighbor in $\train$ is different.
In the closed-world or transductive setting, the search for the $k$-nearest
neighbors does not include the searched-for point $\vx$; in other words, the
nearest neighbors graph does not include self-loops.
In the open-world setting, it is not known if $\vx'$ is contained in
the reference set $\train$, and therefore all points in $\train$ are included in
the $k$-nearest neighbors search.
All of the referenced methods are described in a closed-world setting and do
not explicitly state how to perform inductive outlier detection, yet commonly
used toolkits for (distance-based) outlier detection focus on the open-world
setting \cite{zhaoPyODPythonToolbox2019,muhrOutlierDetectionJlModular2022},
with no clear guidelines on how to transfer the transductive tasks to inductive
tasks.
In Section \ref{sec:probabilistic}, we describe our method in the transductive,
closed-world, and inductive, open-world setting.

\section{Outlier Score
  Normalization} \label{sec:normalization}

As mentioned in the introduction, the outlier scores resulting from
distance-based approaches differ widely in their meaning and are challenging to
interpret.
In some cases, even within a data set, the scores for two different
observations can denote different degrees of outlierness, depending on
different local data distributions, a core motivation for local outlier
detection methods.
Some distance-based methods provide probability estimates, for example,
\cite{kriegelLoOPLocalOutlier2009, kriegelOutlierDetectionArbitrarily2012,
	janssensStochasticOutlierSelection2012, vansteinLocalSubspacebasedOutlier2016},
but these probabilistic interpretations are a core part of the underlying
algorithms and cannot easily be transferred to other algorithmic approaches.
\cite{lateckiOutlierDetectionKernel2007} developed an outlier detection model
based on local kernel density estimates.
\cite{schubertGeneralizedOutlierDetection2014} more generally analyze the
connection of density-based outlier detection algorithms, such as the local
distance-based methods, to kernel density estimation.
The authors show that distance-based density estimation is closely related to
kernel density estimation and claim that local outlier detection methods use
heuristics to determine, perhaps coincidentally, something similar to a
statistical kernel for density estimation.

Instead of algorithm-specific probabilistic interpretations, some authors
propose outlier score normalization schemes independent of the underlying
algorithm.
A simple way of bringing outlier scores to a common scale is to apply a linear
transformation as defined in Equation \ref{eq:norm}, such that the minimum score
is mapped to 0 and the maximum score is mapped to 1.
However, such a min-max scaling approach does not yield a useful probabilistic
interpretation.
\cite{gaoConvertingOutputScores2006} propose two approaches to model outlier
scores as probabilities.
In the first approach, they assume that the posterior probabilities follow a
logistic sigmoid function.
In the second approach, they assume that the outlier scores follow a mixture of
exponential and Gaussian distributions.
In both cases, the authors propose to use an expectation maximization approach
to learn the parameters.
\cite{kriegelInterpretingUnifyingOutlier2011} point out that expectation
maximization approaches to score normalization often converge to a
\enquote{no outliers} or \enquote{all outlier} solution and, instead, propose
to use the cumulative distribution function of a Gaussian or Gamma distribution
to normalize the scores.
Additionally, the authors show the usefulness of post-processing techniques to
ensure an expected value of 0 for normal data points or to increase the
contrast between normal and outlier data points.
\cite{schubertEvaluationOutlierRankings2012} note that a rank-based
normalization can be useful if little knowledge available about the
actual scores and score distributions.

\subsection{Interpretability, Explanation, and Trustworthiness}
\label{sec:explanation}

The interpretability of outlier predictions should not be confused with the
explanation of outlier detection models or the trustworthiness of predictions;
therefore, for the ongoing discussion, we differentiate the terms as follows
and describe them in detail following our differentiation.

\begin{itemize}
	\item \textit{Interpretability}:
	      The ability to judge the relevance of a prediction.
	      % What does the prediction signal?
	\item \textit{Explanation}:
	      The ability to explain the reasoning behind a prediction.
	      % Why does the algorithm determine the prediction?
	\item \textit{Trustworthiness}
	      The ability to describe the confidence behind a prediction.
	      % How confident is the algorithm about its prediction?
\end{itemize}

Explanation is sometimes also referred to as interpretation, but this kind of
interpretation is separate from interpretability.
Explanation algorithms reveal how models make decisions, but
interpretability refers to the intrinsic property in which degree an inference
result is understandable to human beings
\cite{liInterpretableDeepLearning2022}.
There is a growing interest in methods for deriving explanations of outliers,
that is, \enquote{[...] to give the users of some outlier detection method further aid in understanding and evaluating the result with respect to their domain.}
\cite{zimekThereBackAgain2018}.
Explanations highlight why a specific outlier detection model reaches a
particular prediction.
A common approach to explain outlier predictions is to compare normal data
points and outliers in attribute subspaces in which the given outliers show
separability from the normal data
\cite{micenkovaExplainingOutliersSubspace2013,vinhDiscoveringOutlyingAspects2016,machaExplainingAnomaliesGroups2018}.
Other authors derive explanations from statistical models of the normal and
outlier data using minimum distance estimation
\cite{angiulliDiscoveringCharacterizationsBehavior2013}.
The explanation of learning methods and outlier detection methods is
discussed extensively in a research field known as Explainable Artificial
Intelligence, or XAI \cite{samekExplainableAIInterpreting2019}.
Explanations can uncover hidden weaknesses of a model, also known as
\enquote{Clever Hanses} \cite{lapuschkinUnmaskingCleverHans2019}.
The Clever Hans Effect occurs when the learned model produces correct
predictions based on the \enquote{wrong} features, which appears to be a
widespread problem in outlier detection \cite{kauffmannCleverHansEffect2020}.
Another critical aspect of outlier detection predictions is trustworthiness.
Trustworthiness describes an understanding of when a prediction should or should
not be trusted
\cite{leeTrustAutomationDesigning2004,jiangTrustNotTrust2018,ovadiaCanYouTrust2019}.
To achieve better trustworthiness in outlier detection predictions,
\cite{periniQuantifyingConfidenceAnomaly2021} propose to use a Bayesian
approach to add probabilistic uncertainty estimates to outlier scores,
enabling the detector to assign a confidence score to each prediction, which
captures its uncertainty in that prediction.

\section{Probabilistic Outlier Scores}
\label{sec:probabilistic}

In this section, we derive a generic scheme to transform distance-based outlier
detection scores into interpretable outlier scores based on distance
probabilities.
In the generic score normalization approaches mentioned in Section
\ref{sec:explanation}, only the actual scores are used for normalization.
Conversely, the algorithm-specific normalization schemes are generally not
easily transferrable to other algorithms.
A common theme across a vast majority of distance-based outlier detection
methods is the determination of nearest neighbors relationships between data
points.
Determining exact nearest neighbors relationships typically utilizes the
computation of all distance relationships between data points resulting in a
distance matrix $\mM$.
Additionally, it has been shown that brute-force distance computation is
preferable to index methods except for low-dimensional similarity search
problems \cite{muhrHybridCPUGPU2022}.
We note that approximate nearest neighbors approaches are also used for
distance-based outlier detection strategies
\cite{kirnerGoodBadNeighborhood2017}, but this represents a small minority of
methods and is not the focus of our study.
In the closed-world setting, the distance matrix corresponds to a square matrix
of $\sR^{n \times n}$ values for $n$ points in the dataset.
In the open-world setting, the distance matrix between $n$ reference points
$\sR^{n \times n}$ has to be differentiated from the distance matrix
of $m$ query points to $n$ reference points
$\sR^{m \times n}$.
For a point $\vx_i$ and a point $\vx_j$, a value $v_{i,j}$ in the distance
matrix corresponds to the distance $d(\vx_i, \vx_j)$.
%The $k$-nearest neighbors can be described by matrices $\mN \in \sR^{n \times k}$ and $\mN_{\mathrm{test}} \in \sR^{m \times k}$.
Most distance-based approaches use the $k$-nearest neighbors as a context set
to determine the outlier score \cite{schubertLocalOutlierDetection2014}, and any
probabilistic estimate of those scores would be based on the limited
information present in the context set.
In contrast to previous approaches, we assume that the additional information
contained in the distance matrix is useful for normalization.
More concretely, we hypothesize that the additional information can be used to
transform outlier scores into interpretable probabilistic estimates.

Based on a distance matrix of reference points, we define the concept of a
\textit{normalization set}.
A normalization set describes a subset of the distance matrix used for the
probabilistic score normalization.
In the simplest case the entire distance matrix is used as a baseline
normalization set; hence, the normalization set is defined as the distances
contained in the distance matrix $\mM$ between all reference points excluding
self-loops in the matrix diagonal.
Additionally, if the distance measure is symmetric, the normalization set from
the distance matrix can be reduced to its upper or lower triangular set of
values.

\begin{figure}[h]
	\centering
	\includegraphics[width=\linewidth]{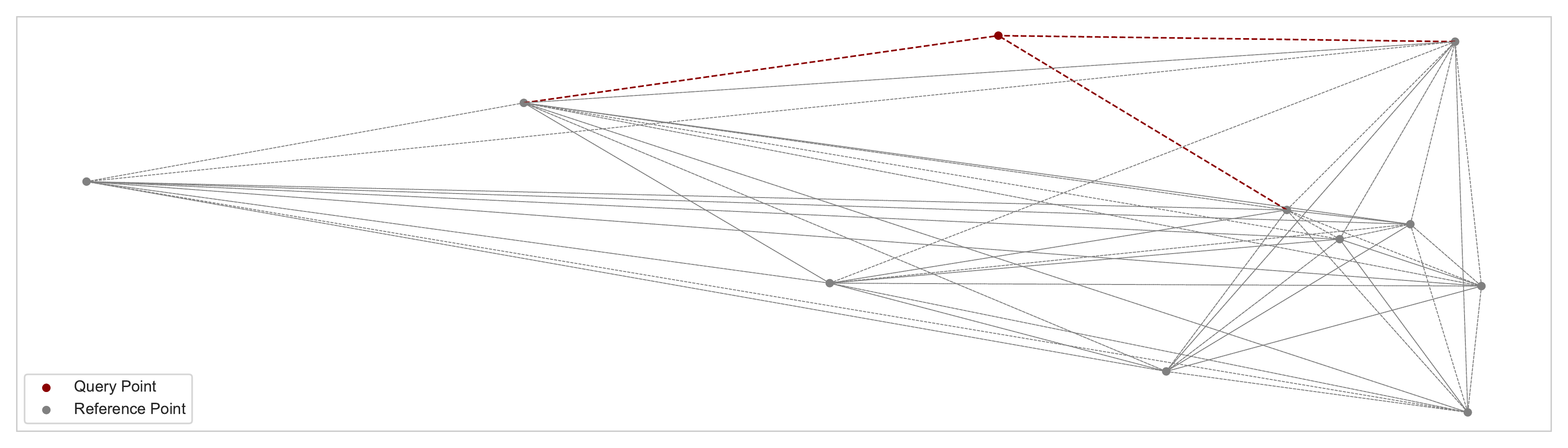}
	\caption{ Visualization of the reference points as a normalization set,
	where the dotted blue lines indicate the connection of the query point to
	its nearest neighbors in the reference set, and the gray lines indicate the
	distance relationships between the reference points.}
	\label{fig:normalization}
\end{figure}

We propose using the normalization set to determine a distance
probability distribution.
For example, in the parametric case, we estimate the parameters of a
distribution $P$ based on the distances in the normalization set.
The distribution of distances has been investigated in the context of feature
similarity \cite{burghoutsDistributionFamilySimilarity2007}, hubness reduction
\cite{schnitzerLocalGlobalScaling2012}, local intrinsic dimensionality
\cite{houleDimensionalityDiscriminabilityDensity2013} or compact sets
\cite{lelloucheDistributionDistancesElements2020}.
\cite{pekalskaClassifiersDissimilaritybasedPattern2000} show, based on the
central limit theorem, that distances are approximately normally distributed
for independent and identically distributed feature vectors.
Under the assumption that the distances in the normalization set follow an
unknown continuous probability distribution, we define a random variable $\rr
	\sim P$ that describes the normalization set.
Given a probability density function $p$ on $\rr$, any distance $d(\vx, \vx')$
can be interpreted as the \textit{distance density} between $\vx$ and
$\vx'$ denoted $p(d(\vx, \vx'))$ or, in short, $p(\vx, \vx')$.
The cumulative \textit{distance distribution}
$f(\vx, \vx') := P(\rr \leq d(\vx, \vx'))$ describes the probability of a distance
in the normalization set being smaller or equal to $d(\vx, \vx')$.
A query point with some distance-based outlier score can be directly
interpreted using its distance distribution, where a probability
of 99\% means that the distance is in the Top-1\% of distances
observed in the normalization set.

In summary, we hypothesize that it is possible to transform distances to
interpretable distance distributions without adverse effects on detection
performance.
We use the Receiver Operating Characteristic (ROC) Area Under the Curve (AUC)
to measure outlier detection performance.
A perfect ranking results in a ROC AUC value of 1, whereas an inverted
perfect ranking would result in a value of 0.
A value of $1/2$ can be interpreted as random guessing
\cite{camposEvaluationUnsupervisedOutlier2016}.
To investigate our assumption, we test our approach in tabular datasets
\cite{camposEvaluationUnsupervisedOutlier2016, muhrOutlierAnomalyDetection2022},
and a common image-based outlier detection dataset
\cite{bergmannMVTecADComprehensive2019, bergmannMVTecAnomalyDetection2021}.

\section{Results} \label{sec:results}

To evaluate the probabilistic transformation on tabular data, we use the
proposed weighted $k$-nearest neighbor approach ($\kNNW$).
The datasets used stem from the DAMI
\cite{camposEvaluationUnsupervisedOutlier2016}
library and the UTSD single-concept benchmark
\cite{muhrOutlierAnomalyDetection2022}.
For both dataset collections, we only use the datasets with five percent of
outliers, each consisting of ten randomly sampled variants, resulting in the
dataset list shown in Table \ref{tab:datasets}.
For DAMI, we use the normalized and deduplicated variants, and for UTSD,
we pre-process the data points with min-max scaling.
We use two-fold, stratified cross-validation to determine a ROC AUC estimate of
the resulting distance-based and probabilistic estimates.
We use Euclidean distance for all evaluations and fix the hyperparameters for
the weighting schemes as $s = a = b = 1$, as shown in Figure \ref{fig:weighting}.
For the number of neighbors $k$ of our tested $\kNNW$ method, we pick the best
parameter from all possible values of $k \in [1,2,\ldots, 100]$.

\begin{table}
	\tiny
	\caption{The datasets used for $k$-nearest neighbors evaluation, where $N$
	denotes the number of samples, $O$ the number of outliers, and $d$ the
	dimensionality of the dataset.}
	\label{tab:datasets}
	\begin{tabular*}{\textwidth}{@{\extracolsep{\fill}}lrrrrr@{}}
	  Dataset & $N$ & $O$ & $d$ & Source & Ref. \\
      \toprule
	  %\colrule
	  Annthyroid               & 6942       & 347        & 21         & ELKI            & \cite{duaUCIMachineLearning2017}                      \\
	  Arrhythmia               & 256        & 12         & 259        & ELKI            & \cite{duaUCIMachineLearning2017}                      \\
	  Cardiotocography         & 1734       & 86         & 21         & ELKI            & \cite{duaUCIMachineLearning2017}                      \\
	  CinCECGTorso             & 373        & 18         & 1639       & UTSD            & \cite{goldbergerPhysioBankPhysioToolkitPhysioNet2000} \\
	  Crop                     & 1052       & 52         & 46         & UTSD            & \cite{tanIndexingClassifyingGigabytes2017}            \\
	  Earthquakes              & 387        & 19         & 512        & UTSD            & \cite{dauUCRTimeSeries2019}                           \\
	  ECG5000                  & 3072       & 153        & 140        & UTSD            & \cite{goldbergerPhysioBankPhysioToolkitPhysioNet2000} \\
	  ECGFiveDays              & 465        & 23         & 136        & UTSD            & \cite{dauUCRTimeSeries2019}                           \\
	  ElectricDevices          & 4500       & 225        & 96         & UTSD            & \cite{dauUCRTimeSeries2019}                           \\
	  FaceAll                  & 344        & 17         & 131        & UTSD            & \cite{dauUCRTimeSeries2019}                           \\
	  FordA                    & 2660       & 133        & 500        & UTSD            & \cite{dauUCRTimeSeries2019}                           \\
	  FordB                    & 2380       & 119        & 500        & UTSD            & \cite{dauUCRTimeSeries2019}                           \\
	  FreezerRegularTrain      & 1578       & 78         & 301        & UTSD            & \cite{murrayDataManagementPlatform2015}               \\
	  HandOutlines             & 921        & 46         & 2709       & UTSD            & \cite{davisPredictiveModellingBone2013}               \\
	  HeartDisease             & 157        & 7          & 13         & ELKI            & \cite{duaUCIMachineLearning2017}                      \\
	  Hepatitis                & 70         & 3          & 19         & ELKI            & \cite{duaUCIMachineLearning2017}                      \\
	  InternetAds              & 1682       & 84         & 1555       & ELKI            & \cite{duaUCIMachineLearning2017}                      \\
	  ItalyPowerDemand         & 575        & 28         & 24         & UTSD            & \cite{keoghIntelligentIconsIntegrating2006}           \\
	  MedicalImages            & 625        & 31         & 99         & UTSD            & \cite{dauUCRTimeSeries2019}                           \\
	  MixedShapesRegularTrain  & 793        & 39         & 1024       & UTSD            & \cite{wangAnnotatingHistoricalArchives2010}           \\
	  MoteStrain               & 721        & 36         & 84         & UTSD            & \cite{sunOnlineLatentVariable2005}                    \\
	  PageBlocks               & 5139       & 256        & 10         & ELKI            & \cite{duaUCIMachineLearning2017}                      \\
	  Parkinson                & 50         & 2          & 22         & ELKI            & \cite{duaUCIMachineLearning2017}                      \\
	  PhalangesOutlinesCorrect & 1787       & 89         & 80         & UTSD            & \cite{davisPredictiveModellingBone2013}               \\
	  Pima                     & 526        & 26         & 8          & ELKI            & \cite{duaUCIMachineLearning2017}                      \\
	  SemgHandGenderCh2        & 568        & 28         & 1500       & UTSD            & \cite{sapsanisImprovingEMGBased2013}                  \\
	  SonyAIBORobotSurface2    & 635        & 31         & 65         & UTSD            & \cite{mueenLogicalshapeletsExpressivePrimitive2011}   \\
	  SpamBase                 & 2661       & 133        & 57         & ELKI            & \cite{duaUCIMachineLearning2017}                      \\
	  Stamps                   & 325        & 16         & 9          & ELKI            & \cite{micenkovaStampVerificationAutomated2015}        \\
	  StarLightCurves          & 5607       & 280        & 1024       & UTSD            & \cite{rebbapragadaFindingAnomalousPeriodic2009}      \\
	  Strawberry               & 369        & 18         & 235        & UTSD            & \cite{dauUCRTimeSeries2019}                           \\
	  TwoLeadECG               & 611        & 30         & 82         & UTSD            & \cite{goldbergerPhysioBankPhysioToolkitPhysioNet2000} \\
	  UWaveGestureLibraryAll   & 589        & 29         & 945        & UTSD            & \cite{liuUWaveAccelerometerbasedPersonalized2009}     \\
	  Wafer                    & 6738       & 336        & 152        & UTSD            & \cite{olszewskiGeneralizedFeatureExtraction2001}      \\
	  Yoga                     & 1863       & 93         & 426        & UTSD            & \cite{dauUCRTimeSeries2019}							\\
	  \botrule
	\end{tabular*}
\end{table}

\begin{figure}
	\centering
	\includegraphics[width=\linewidth]{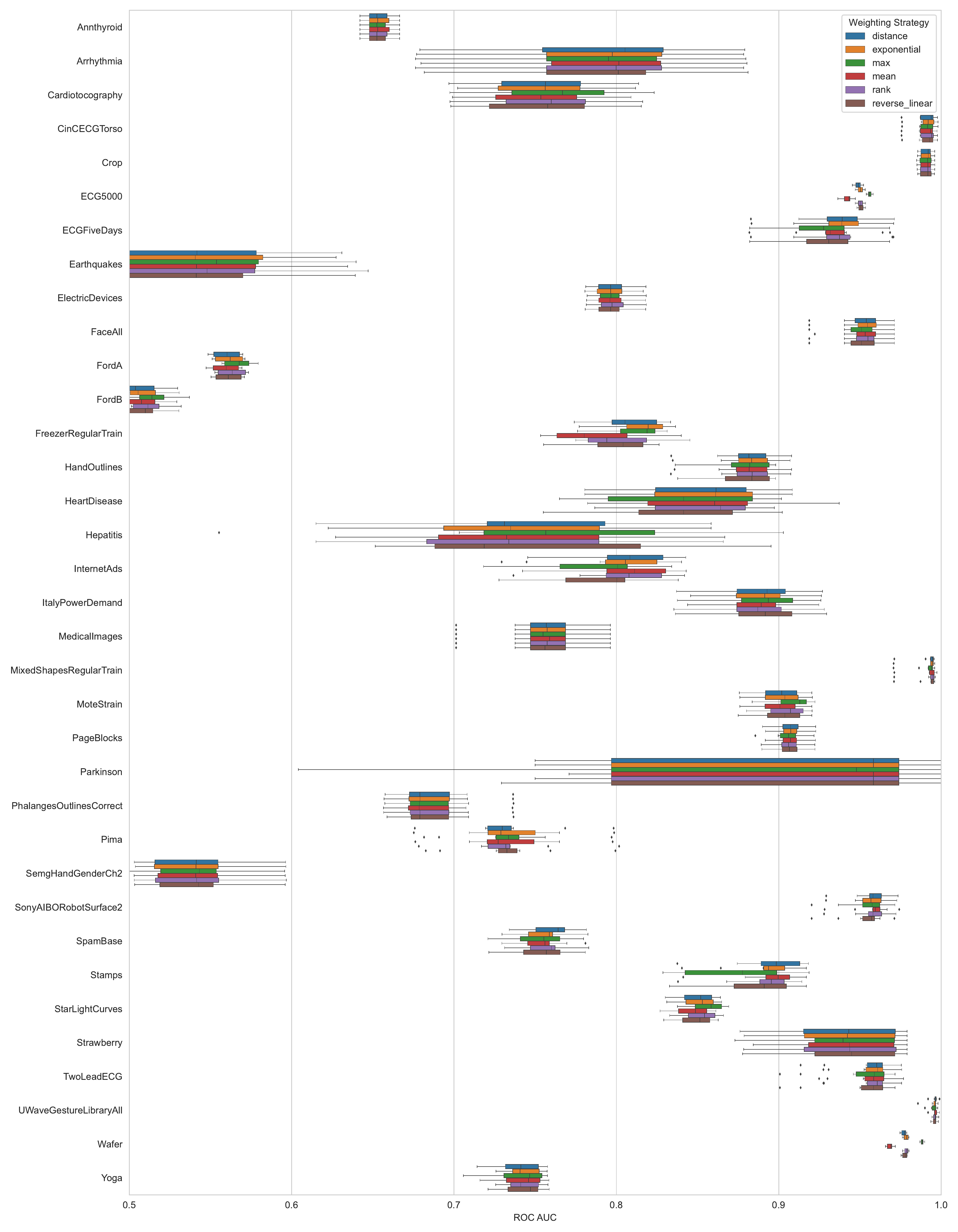}
	\caption{ ROC AUC results of the different weighting schemes for each
	dataset over all examined distributions. }
	\label{fig:results_weighting}
\end{figure}

\begin{figure}
	\centering
	\includegraphics[width=\linewidth]{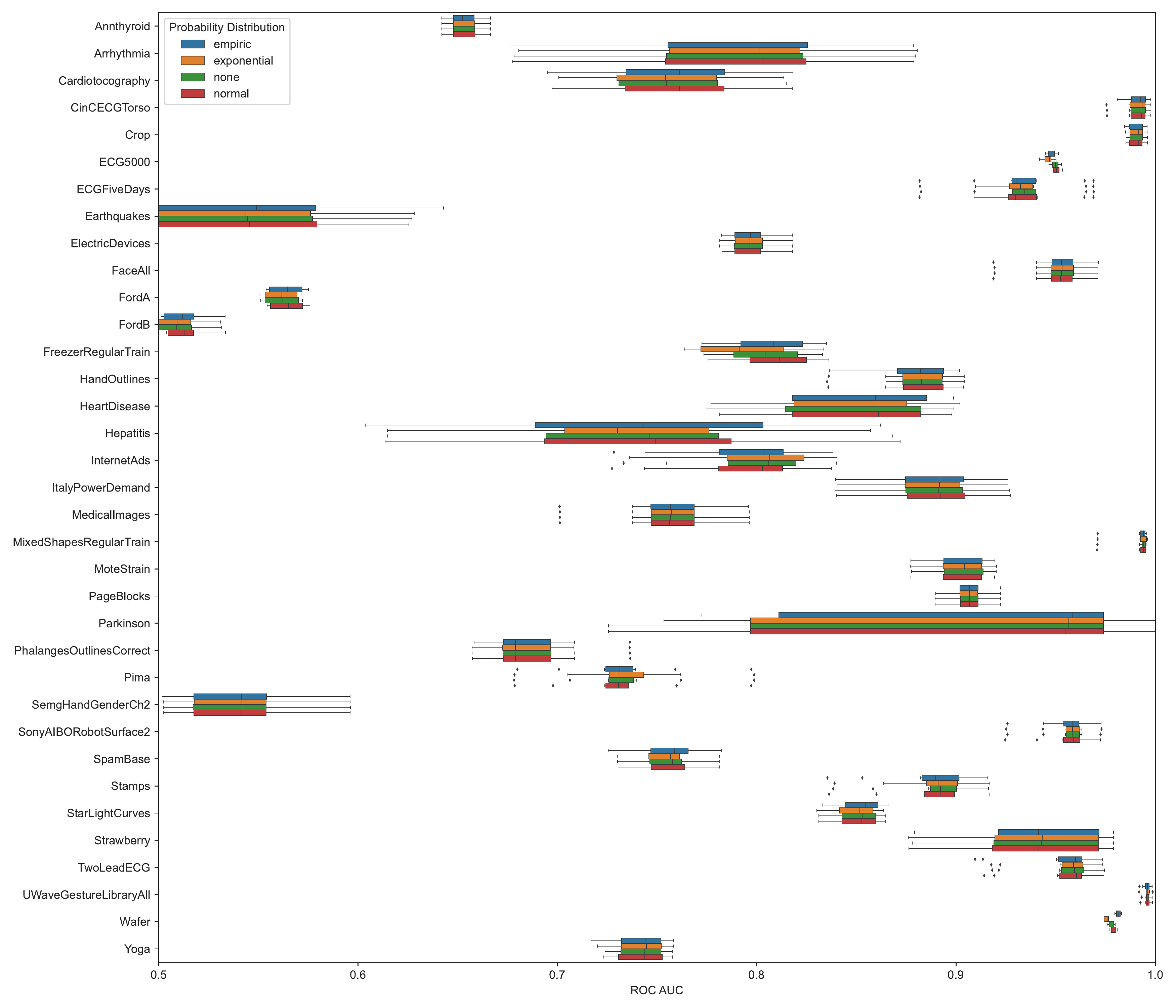}
	\caption{ ROC AUC results of different probability distributions for each
	dataset over all weighting schemes, where 'none' denotes the original
	scores without normalization. }
	\label{fig:results_distribution}
\end{figure}

\begin{figure}
	\centering
	\includegraphics[width=\linewidth]{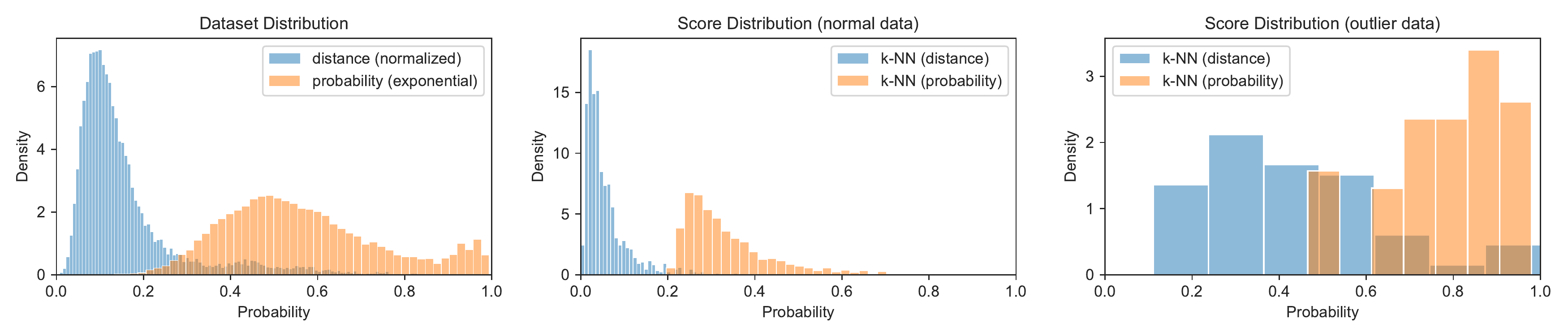}
	\caption{ Transformation of distances to an exponential cumulative distance
	distribution for the first variant of the Crop dataset. }
	\label{fig:crop}
\end{figure}

In our first analysis, we compare the performance of different weighting
schemes over all described datasets.
We find that, over all datasets, there is no difference in predictive
performance between the different weighting schemes.
There can be more considerable weighting scheme differences for individual
datasets; however, those are dataset-specific and must be investigated
case-by-case, as shown in Figure \ref{fig:results_weighting}.
We also note that weighting-scheme hyperparameter optimization might yield
additional improvements, which we did not address in our analysis.
In our second analysis, we investigate the impact of score normalization using
different probability distributions.
We examine a normal, exponential, and empirical distribution and compare it to
the base case where no distribution is used for normalization.
Because the cumulative distribution functions are monotonically non-decreasing,
the ranking should be stable after the transformation, but due to the
limited precision of the computations, it is not guaranteed that the
transformation is ranking stable.
In Figure \ref{fig:results_distribution}, we show that the ROC AUC result after
the transformation matches the original result and the transformation is indeed
ranking stable.
From an interpretability perspective, there are datasets where using the entire
distance matrix as a normalization set results in useful probabilistic
estimates, for example, the Crop dataset shown in Figure \ref{fig:crop}.

However, using the entire distance matrix for normalization often leads to
low probabilistic estimates for normal and outlier data points.
The reason is that the resulting weighted neighbor distance is consistently low
compared to all other distances in the dataset, even for outliers.
In this case, a different normalization set has to be extracted from the
distance matrix; for example, the $m$-neighborhood consisting of the distances
to the $m$-closest reference points.

\begin{figure}
	\centering
	\includegraphics[width=\linewidth]{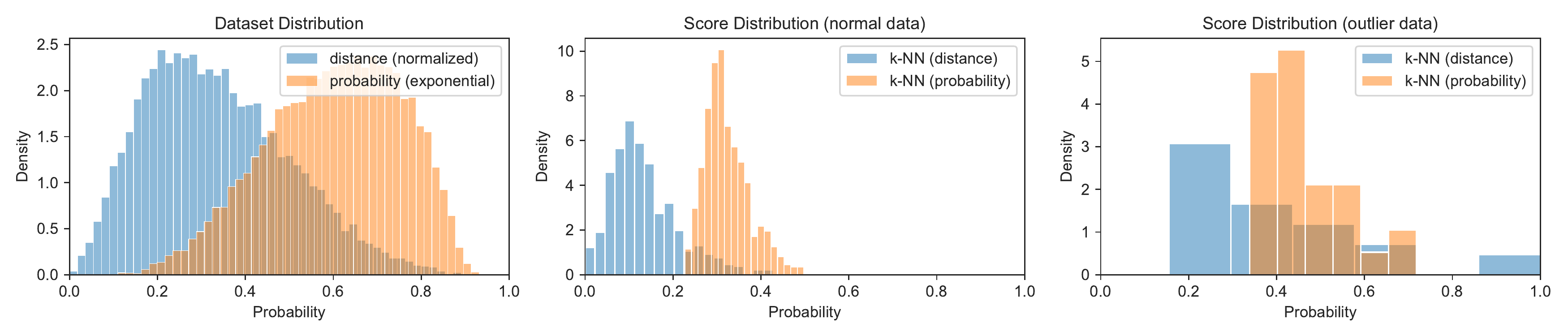}
	\caption{ Transformation of distances to an exponential distance
	distribution for the first variant of the TwoLeadECG dataset showing low
	outlier probability. }
	\label{fig:ecg-ref}
\end{figure}

\begin{figure}
	\centering
	\includegraphics[width=\linewidth]{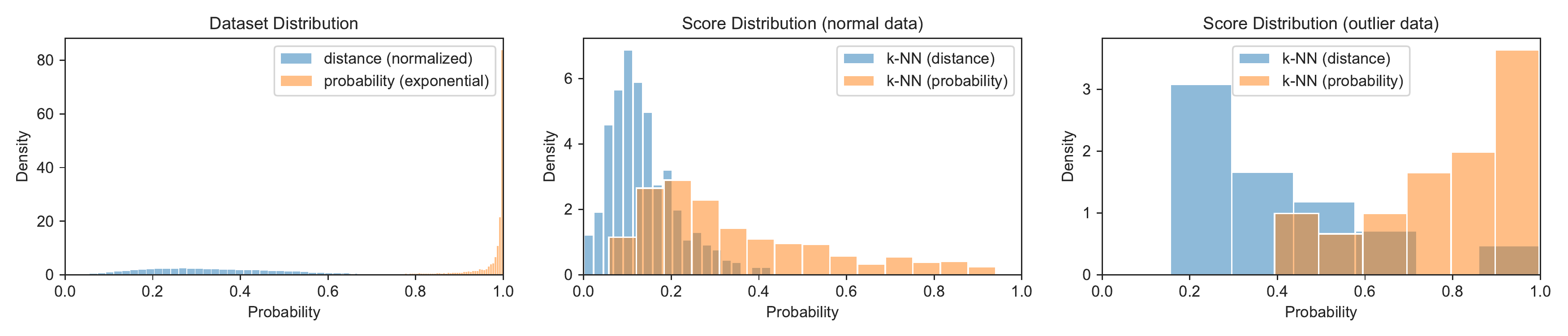}
	\caption{ Exponential distance distribution using the 99 element neighborhood
	as a normalization set for the first variant of the TwoLeadECG dataset
	to achieve a suitable cut-off threshold and increased contrast. }
	\label{fig:ecg-nn}
\end{figure}

\begin{figure}
	\centering
	\includegraphics[width=\linewidth]{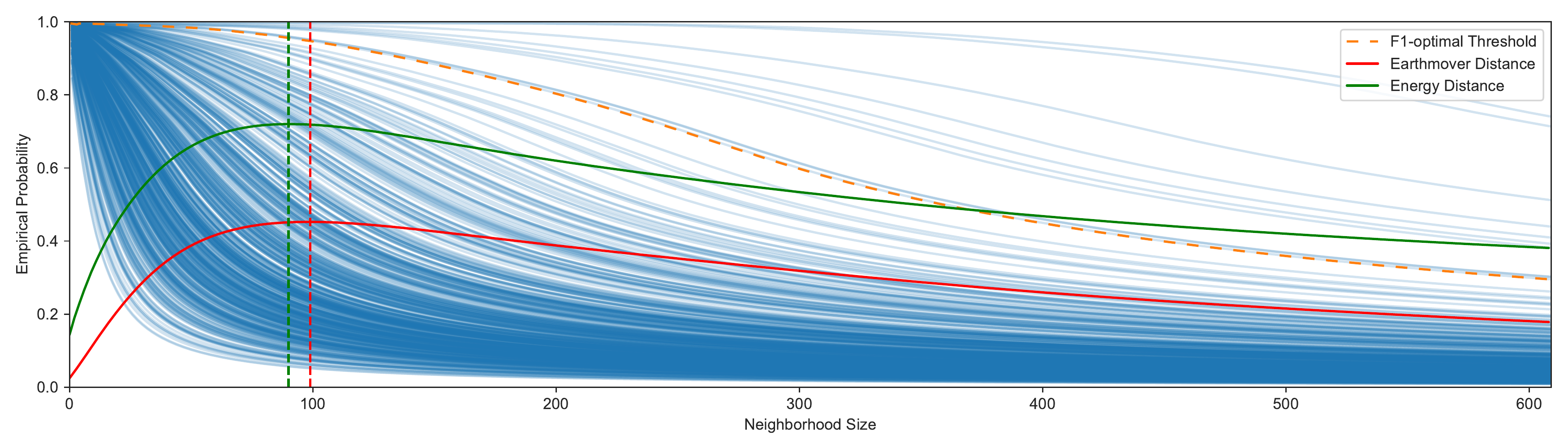}
	\caption{ Each blue line refers to the probabilistic outlier score of the
	first TwoLeadECG dataset variant.
	The green and red lines show the statistical distance between the normal
	score distribution and the outlier score distribution, with the dashed
	vertical lines depicting the corresponding maximum contrast value.
	The orange dashed line shows the F1-score optimal cut-off.
	}
	\label{fig:cutoff}
\end{figure}

It is possible to analyze multiple normalization sets for a single prediction
to provide more context for interpretation.
For example, a prediction can be interpreted using different neighborhood
probabilities from $m \in [1,2,\ldots,200]$ to determine an appropriate cut-off
threshold.
The cut-off decision always relies on the data characteristics and domain
knowledge.
In Figure \ref{fig:cutoff}, we plot the optimal cut-off threshold for different
neighborhood sizes, but note that this optimal threshold is often difficult
to determine.
To evaluate an optimal cut-off threshold, it is necessary to evaluate it
against a performance metric such as the F1-score requiring normal and
outlier labeled data, which is often unavailable.
However, using a probabilistic neighborhood analysis as shown in Figure
\ref{fig:cutoff} drastically simplifies the identification of a cut-off value,
even when labels are unavailable.
Thus, in addition to the improved interpretability, choosing an appropriate
normalization set allows for a flexible definition of a cut-off threshold to
transform outlier scores into class labels.
Furthermore, it is possible to increase the contrast between normal and outlier
data points using the right normalization set and distribution.
Using statistical distances as a measure of contrast between normal and outlier
scores, we can identify an optimal normalization set size.
To give an example for the TwoLeadECG dataset, the statistical measures of
contrast result in a contrast-optimal neighborhood size of $m=90$ or $m=99$
depending on the statistical distance used, with a cut-off threshold of
approximately 95\%, as shown in Figure \ref{fig:cutoff}.
In Figure \ref{fig:ecg-ref} and Figure \ref{fig:ecg-nn}, we compare
the initial probabilistic estimates using the entire distance matrix to the
smaller, local normalization set identified in Figure \ref{fig:cutoff} and
clearly demonstrate the increased contrast using a smaller normalization set.

\FloatBarrier

For image-based datasets, we extend the PatchCore methodology
\cite{rothTotalRecallIndustrial2022} to \textit{ProbabilisticPatchCore}.
We evaluate the model on the datasets provided by MVTecAD, as shown in
Table \ref{tab:mvtecad}.
A major difference between tabular $k$-nearest neighbors outlier detection and
image outlier detection is that the image models may result in pixel-wise and
image-wise outlier scores.
In the pixel-wise case, each pixel of an observation is scored and in the
image-wise case a single score is obtained for the entire image.

\begin{table}
	\tiny
	\caption{MVTecAD \cite{bergmannMVTecAnomalyDetection2021} image datasets
	for the evaluation of ProbabilisticPatchCore.}
	\label{tab:mvtecad}
	\begin{tabular*}{\textwidth}{@{\extracolsep{\fill}}lrrrrrr@{}}
		Dataset    & Train (normal) & Test (normal) & Test (outlier) & Masks & Groups & Shape              \\
		Carpet     & 280            & 28            & 89             & 97    & 5      & $1024 \times 1024$ \\
		Grid       & 264            & 21            & 57             & 170   & 5      & $1024 \times 1024$ \\
		Leather    & 245            & 32            & 92             & 99    & 5      & $1024 \times 1024$ \\
		Tile       & 230            & 33            & 84             & 86    & 5      & $840 \times 840$   \\
		Wood       & 247            & 19            & 60             & 168   & 5      & $1024 \times 1024$ \\
		Bottle     & 209            & 20            & 63             & 68    & 3      & $900 \times 900$   \\
		Cable      & 224            & 58            & 92             & 151   & 8      & $1024 \times 1024$ \\
		Capsule    & 219            & 23            & 109            & 114   & 5      & $1000 \times 1000$ \\
		Hazelnut   & 391            & 40            & 70             & 136   & 4      & $1024 \times 1024$ \\
		Metal Nut  & 220            & 22            & 93             & 132   & 4      & $700 \times 700$   \\
		Pill       & 267            & 26            & 141            & 245   & 7      & $800 \times 800$   \\
		Screw      & 320            & 41            & 119            & 135   & 5      & $1024 \times 1024$ \\
		Toothbrush & 60             & 12            & 30             & 66    & 1      & $1024 \times 1024$ \\
		Transistor & 213            & 60            & 40             & 44    & 4      & $1024 \times 1024$ \\
		Zipper     & 240            & 32            & 119            & 177   & 7      & $1024 \times 1024$ \\
	  \botrule
	\end{tabular*}
\end{table}

\begin{figure}
	\centering
	\includegraphics[width=\linewidth]{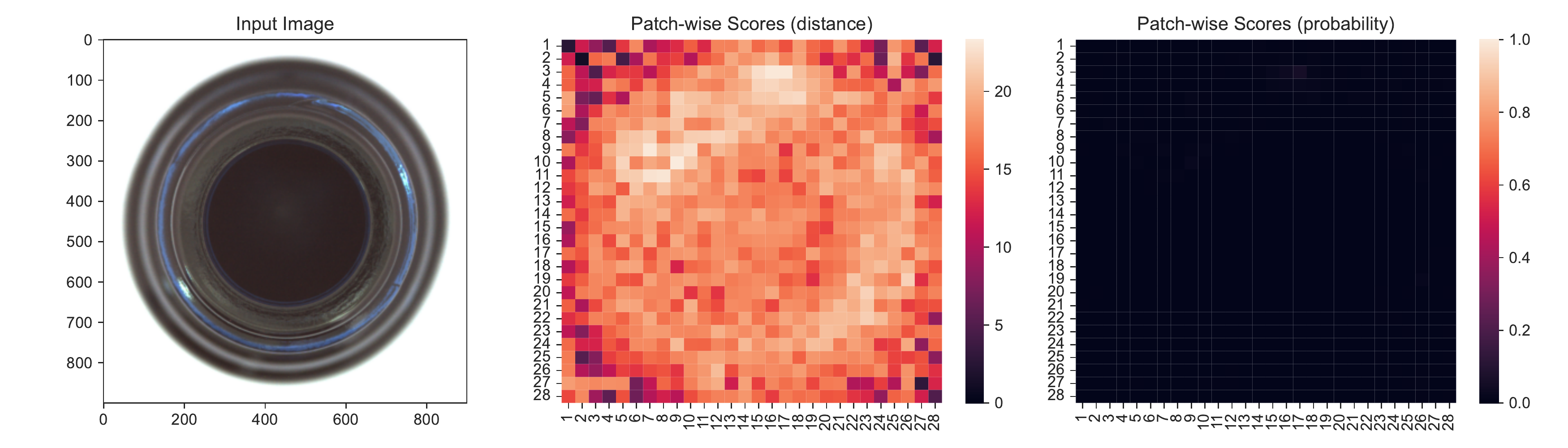}
	\caption{ Patch-wise scores of a normal sample of the Bottle
	dataset showing interpretability differences. }
	\label{fig:bottle-normal}
\end{figure}

\begin{figure}
	\centering
	\includegraphics[width=\linewidth]{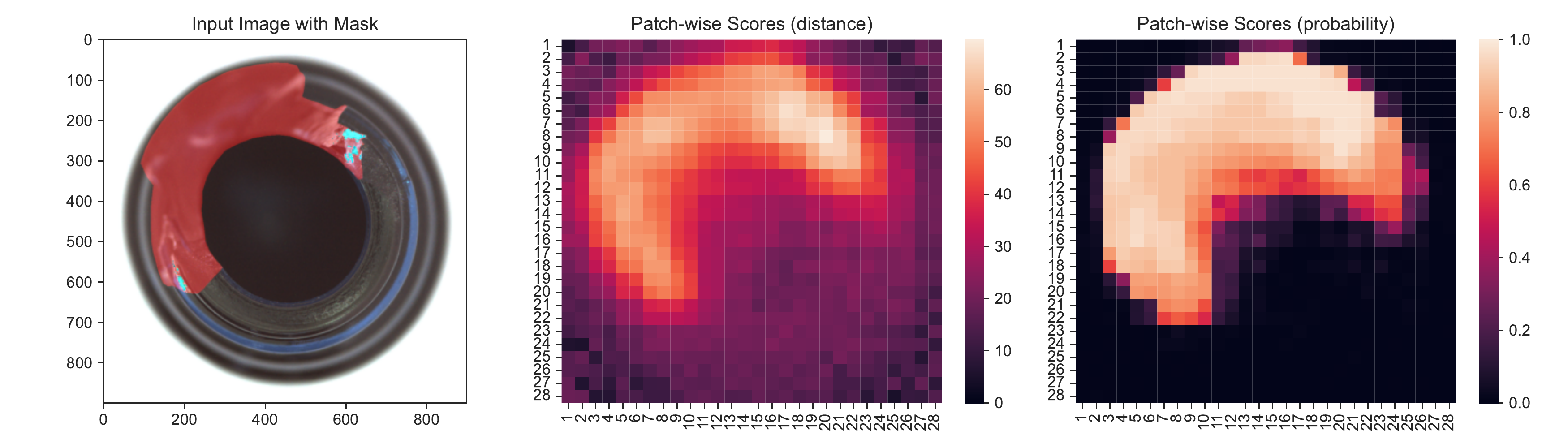}
	\caption{ Patch-wise scores of an outlier sample of the Bottle
	dataset exhibiting increased contrast. }
	\label{fig:bottle-outlier}
\end{figure}

The PatchCore model is similar to $\kthNN$, but uses a core-set sampled memory
bank of patch-wise feature vectors that are generated using a pre-trained
neural network.
For PatchCore, the pixel-wise scores are determined through interpolation of
the patch-wise scores; thus, it is not necessary to estimate a distance
distribution per pixel, but one distribution per patch.
Like the authors, we use the second and third layer of a \textit{WideResNet50}
trained on ImageNet \cite{dengImageNetLargescaleHierarchical2009}
to determine $28 \times 28$ patches.
% TODO: mention exact parameters used
To transform the pixel-wise scores into probabilistic estimates, we learn
a patch-wise distributions and transform the scores to probabilistic estimates
accordingly.
In Figure \ref{fig:bottle-normal}, we highlight the challenge of
interpretability based on a normal data sample; without additional context,
it is not clear how to interpret the resulting distance-based scores.
In addition to the improved interpretability, we find that the probabilistic
normalization greatly increases the contrast between normal and outlier data
points in the image detection tasks as visible in Figure
\ref{fig:bottle-outlier}.

\FloatBarrier
\section{Conclusion}
\label{sec:conclusion}

% What are the contributions to the field? Summary of key findings.
We show that it is possible to transform distance-based outlier scores into
interpretable probabilistic estimates.
To demonstrate the viability of our approach, we derive and test a generalized,
weighted $k$-nearest neighbors outlier detection model on a several tabular
datasets and a probabilistic PatchCore model on image datasets.
We show that the resulting probabilistic scores increase the contrast between
normal and outlier data points and can easily be added to existing
distance-based outlier detection methods.
% What is new in your work? Strengths of the work.
In contrast to previous score normalization techniques, which use solely the
information contained in the outlier scores to derive a normalization,
we make use of the distances to other data points as an additional source of
information for normalization.
Another interesting aspect of our analysis is showing that the probabilistic
transformation increases the contrast between normal and outlier points, which
should be further explored.
Specifically, we find that there might be an optimal normalization set that
maximially increases the contrast between normal and outlier points and future
research is necessary to define measures of contrast and methods to
identify an optimal normalization set for a given contrast measure.
Because distance-based outlier detection techniques rely on distance
computations for nearest neighbors search, our approach can be applied to
a wide range of detection techniques.
% What are the current limitations? Results of critical analysis.
In our experiments, we use the common Euclidean distance metric, but other,
possibly non-metric, distance measures are also used for outlier detection,
and should also be investigated using our probabilistic score transformation.
We investigated only the most apparent normalization sets, but there may be
various other useful normalization sets hidden in the distances between points.
Another limitation of our examination is the usage of real-world datasets,
which limits the theoretical analysis of our approach, such as the
normalization behaviour under specific dataset distributions.
Our proposed normalization approach should be investigated more thoroughly
in a theoretical setting to identify the limits of our approach and potentially
proof some of the properties observed in our evaluation.
% What to do in the future? Proposal of hypotheses.
Our proposed generalization of weighted nearest neighbors outlier detection
should be analysed in more detail to thoroughly compare weighting strategies
and weighting hyperparameters.
A large body of research investigates sampling and subspacing techniques for
distance-based outlier detection and future researchers should evaluate
the usefulness of probabilistic intepretations for such models.
Another important area of outlier detection research is how to combine different
detection models into ensembles that improve upon the individual models, which
typically necessitates score normalization and, therefore, could benefit
from probabilistic normalization.
We further highlight the importance of a distinction between the open-world and
closed-world setting for distance-based outlier detection and propose
such a distinction for future distance-based methods.

\section*{Conflict of Interest Statement}

The authors declare that the research was conducted in the absence of any
commercial or financial relationships that could be construed as a potential
conflict of interest.

\section*{Author Contributions}

DM devised the ideas presented in this paper and wrote the first draft of the
manuscript.
All authors contributed to manuscript revision, read, and approved the
submitted version.

\section*{Data Availability Statement}
The datasets for this investigation can be found in
\href{https://www.dbs.ifi.lmu.de/research/outlier-evaluation/DAMI/}{DAMI},
\href{https://outlier-detection.github.io/utsd/}{UTSD}, and
\href{https://www.mvtec.com/company/research/datasets/mvtec-ad}{MVTecAD}.
The code to reproduce the experiments can be found on
\href{https://github.com/davnn/probabilistic-distance}{GitHub}.
% Please see the availability of data guidelines for more information, at https://www.frontiersin.org/guidelines/policies-and-publication-ethics#materials-and-data-policies

\bibliographystyle{Frontiers-Harvard} %  Many Frontiers journals use the Harvard referencing system (Author-date), to find the style and resources for the journal you are submitting to: https://zendesk.frontiersin.org/hc/en-us/articles/360017860337-Frontiers-Reference-Styles-by-Journal. For Humanities and Social Sciences articles please include page numbers in the in-text citations 
\bibliography{references}

\providecommand{\noopsort}[1]{}
\begin{thebibliography}{92}
\providecommand{\natexlab}[1]{#1}
\expandafter\ifx\csname urlstyle\endcsname\relax
  \providecommand{\doi}[1]{doi:\discretionary{}{}{}#1}\else
  \providecommand{\doi}{doi:\discretionary{}{}{}\begingroup
  \urlstyle{rm}\Url}\fi
\providecommand{\selectlanguage}[1]{\relax}
\providecommand{\bibAnnoteFile}[1]{%
  \IfFileExists{#1}{\begin{quotation}\noindent\textsc{Key:} #1\\
  \textsc{Annotation:}\ \input{#1}\end{quotation}}{}}
\providecommand{\bibAnnote}[2]{%
  \begin{quotation}\noindent\textsc{Key:} #1\\
  \textsc{Annotation:}\ #2\end{quotation}}

\bibitem[{Aggarwal and Yu(2002)}]{aggarwalOutlierDetectionHigh2002}
Aggarwal, C. and Yu, P. (2002).
\newblock Outlier {{Detection}} for {{High Dimensional Data}}.
\newblock \emph{ACM SIGMOD Record} 30.
\newblock \doi{10.1145/376284.375668}
\bibAnnoteFile{aggarwalOutlierDetectionHigh2002}

\bibitem[{Aggarwal and
  Sathe(2015)}]{aggarwalTheoreticalFoundationsAlgorithms2015}
Aggarwal, C.~C. and Sathe, S. (2015).
\newblock Theoretical {{Foundations}} and {{Algorithms}} for {{Outlier
  Ensembles}}.
\newblock \emph{SIGKDD Explor. Newsl.} 17.
\newblock \doi{10.1145/2830544.2830549}
\bibAnnoteFile{aggarwalTheoreticalFoundationsAlgorithms2015}

\bibitem[{Agrawal(2009)}]{agrawalLocalSubspaceBased2009}
Agrawal, A. (2009).
\newblock Local {{Subspace Based Outlier Detection}}.
\newblock In \emph{Contemporary {{Computing}}}, eds. S.~Ranka, S.~Aluru,
  R.~Buyya, Y.-C. Chung, S.~Dua, A.~Grama, S.~K.~S. Gupta, R.~Kumar, and V.~V.
  Phoha ({Berlin, Heidelberg}: {Springer Berlin Heidelberg}), Communications in
  {{Computer}} and {{Information Science}}.
\newblock \doi{10.1007/978-3-642-03547-0_15}
\bibAnnoteFile{agrawalLocalSubspaceBased2009}

\bibitem[{Alghushairy et~al.(2021)Alghushairy, Alsini, Soule, and
  Ma}]{alghushairyReviewLocalOutlier2021}
Alghushairy, O., Alsini, R., Soule, T., and Ma, X. (2021).
\newblock A {{Review}} of {{Local Outlier Factor Algorithms}} for {{Outlier
  Detection}} in {{Big Data Streams}}.
\newblock \emph{BDCC} 5.
\newblock \doi{10.3390/bdcc5010001}
\bibAnnoteFile{alghushairyReviewLocalOutlier2021}

\bibitem[{Angiulli et~al.(2013)Angiulli, Fassetti, and
  Palopoli}]{angiulliDiscoveringCharacterizationsBehavior2013}
Angiulli, F., Fassetti, F., and Palopoli, L. (2013).
\newblock Discovering {{Characterizations}} of the {{Behavior}} of {{Anomalous
  Subpopulations}}.
\newblock \emph{IEEE Transactions on Knowledge and Data Engineering} 25.
\newblock \doi{10.1109/TKDE.2012.58}
\bibAnnoteFile{angiulliDiscoveringCharacterizationsBehavior2013}

\bibitem[{Angiulli and Pizzuti(2002)}]{angiulliFastOutlierDetection2002}
Angiulli, F. and Pizzuti, C. (2002).
\newblock Fast {{Outlier Detection}} in {{High Dimensional Spaces}}.
\newblock In \emph{Principles of {{Data Mining}} and {{Knowledge Discovery}}},
  eds. G.~Goos, J.~Hartmanis, J.~{\noopsort{leeuwen}}{van Leeuwen}, J.~G.
  Carbonell, J.~Siekmann, T.~Elomaa, H.~Mannila, and H.~Toivonen ({Berlin,
  Heidelberg}: {Springer Berlin Heidelberg}), Lecture {{Notes}} in {{Computer
  Science}}.
\newblock \doi{10.1007/3-540-45681-3_2}
\bibAnnoteFile{angiulliFastOutlierDetection2002}

\bibitem[{Barnett and Lewis(1978)}]{barnettOutliersStatisticalData1978}
Barnett, V. and Lewis, T. (1978).
\newblock \emph{Outliers in {{Statistical Data}}} ({John Wiley \& Sons, Inc})
\bibAnnoteFile{barnettOutliersStatisticalData1978}

\bibitem[{Bergman et~al.(2020)Bergman, Cohen, and
  Hoshen}]{bergmanDeepNearestNeighbor2020}
Bergman, L., Cohen, N., and Hoshen, Y. (2020).
\newblock Deep nearest neighbor anomaly detection.
\newblock \emph{CoRR} abs/2002.10445
\bibAnnoteFile{bergmanDeepNearestNeighbor2020}

\bibitem[{Bergmann et~al.(2021)Bergmann, Batzner, Fauser, Sattlegger, and
  Steger}]{bergmannMVTecAnomalyDetection2021}
Bergmann, P., Batzner, K., Fauser, M., Sattlegger, D., and Steger, C. (2021).
\newblock The {{MVTec Anomaly Detection Dataset}}: {{A Comprehensive Real-World
  Dataset}} for {{Unsupervised Anomaly Detection}}.
\newblock \emph{Int J Comput Vis} 129.
\newblock \doi{10.1007/s11263-020-01400-4}
\bibAnnoteFile{bergmannMVTecAnomalyDetection2021}

\bibitem[{Bergmann et~al.(2019)Bergmann, Fauser, Sattlegger, and
  Steger}]{bergmannMVTecADComprehensive2019}
Bergmann, P., Fauser, M., Sattlegger, D., and Steger, C. (2019).
\newblock {{MVTec AD}} \textemdash{} {{A Comprehensive Real-World Dataset}} for
  {{Unsupervised Anomaly Detection}}.
\newblock In \emph{2019 {{IEEE}}/{{CVF Conference}} on {{Computer Vision}} and
  {{Pattern Recognition}} ({{CVPR}})}.
\newblock \doi{10.1109/CVPR.2019.00982}
\bibAnnoteFile{bergmannMVTecADComprehensive2019}

\bibitem[{Breunig et~al.(2000)Breunig, Kriegel, Ng, and
  Sander}]{breunigLOFIdentifyingDensityBased2000}
Breunig, M.~M., Kriegel, H.-P., Ng, R.~T., and Sander, J. (2000).
\newblock {{LOF}}: {{Identifying Density-Based Local Outliers}}.
\newblock In \emph{Proceedings of the 2000 {{ACM SIGMOD International
  Conference}} on {{Management}} of {{Data}}: 2000, {{Dallas}}, {{Texas}},
  {{United States}}, {{May}} 15- 8, 2000}, eds. M.~Dunham, J.~F. Naughton,
  W.~Chen, and N.~Koudas ({New York, NY, USA}: {Association for Computing
  Machinery}).
\newblock \doi{10.1145/342009.335388}
\bibAnnoteFile{breunigLOFIdentifyingDensityBased2000}

\bibitem[{Burghouts et~al.(2007)Burghouts, Smeulders, and
  Geusebroek}]{burghoutsDistributionFamilySimilarity2007}
Burghouts, G., Smeulders, A., and Geusebroek, J.-m. (2007).
\newblock The {{Distribution Family}} of {{Similarity Distances}}.
\newblock In \emph{Advances in {{Neural Information Processing Systems}}}
  ({Curran Associates, Inc.}), vol.~20
\bibAnnoteFile{burghoutsDistributionFamilySimilarity2007}

\bibitem[{Cabero et~al.(2021)Cabero, Epifanio, Pi{\'e}rola, and
  Ballester}]{caberoArchetypeAnalysisNew2021}
Cabero, I., Epifanio, I., Pi{\'e}rola, A., and Ballester, A. (2021).
\newblock Archetype analysis: {{A}} new subspace outlier detection approach.
\newblock \emph{Knowledge-Based Systems} 217.
\newblock \doi{10.1016/j.knosys.2021.106830}
\bibAnnoteFile{caberoArchetypeAnalysisNew2021}

\bibitem[{Campos et~al.(2016)Campos, Zimek, Sander, Campello, Micenkov{\'a},
  Schubert et~al.}]{camposEvaluationUnsupervisedOutlier2016}
Campos, G.~O., Zimek, A., Sander, J., Campello, R. J. G.~B., Micenkov{\'a}, B.,
  Schubert, E., et~al. (2016).
\newblock On the evaluation of unsupervised outlier detection: Measures,
  datasets, and an empirical study.
\newblock \emph{Data Min Knowl Disc} 30.
\newblock \doi{10.1007/s10618-015-0444-8}
\bibAnnoteFile{camposEvaluationUnsupervisedOutlier2016}

\bibitem[{Chandola et~al.(2009)Chandola, Banerjee, and
  Kumar}]{chandolaAnomalyDetection2009}
Chandola, V., Banerjee, A., and Kumar, V. (2009).
\newblock Anomaly detection.
\newblock \emph{ACM Comput. Surv.} 41.
\newblock \doi{10.1145/1541880.1541882}
\bibAnnoteFile{chandolaAnomalyDetection2009}

\bibitem[{Cohen and Hoshen(2021)}]{cohenSubImageAnomalyDetection2021}
[Dataset] Cohen, N. and Hoshen, Y. (2021).
\newblock Sub-{{Image Anomaly Detection}} with {{Deep Pyramid Correspondences}}
\bibAnnoteFile{cohenSubImageAnomalyDetection2021}

\bibitem[{Dang et~al.(2015)Dang, Ngan, and
  Liu}]{dangDistancebasedKnearestNeighbors2015}
Dang, T.~T., Ngan, H.~Y., and Liu, W. (2015).
\newblock Distance-based k-nearest neighbors outlier detection method in
  large-scale traffic data.
\newblock In \emph{2015 {{IEEE International Conference}} on {{Digital Signal
  Processing}} ({{DSP}})}.
\newblock \doi{10.1109/ICDSP.2015.7251924}
\bibAnnoteFile{dangDistancebasedKnearestNeighbors2015}

\bibitem[{Dau et~al.(2019)Dau, Bagnall, Kamgar, Yeh, Zhu, Gharghabi
  et~al.}]{dauUCRTimeSeries2019}
Dau, H.~A., Bagnall, A., Kamgar, K., Yeh, C.-C.~M., Zhu, Y., Gharghabi, S.,
  et~al. (2019).
\newblock The {{UCR}} time series archive.
\newblock \emph{IEEE/CAA Journal of Automatica Sinica} 6.
\newblock \doi{10.1109/JAS.2019.1911747}
\bibAnnoteFile{dauUCRTimeSeries2019}

\bibitem[{Davis(2013)}]{davisPredictiveModellingBone2013}
Davis, L.~M. (2013).
\newblock \emph{Predictive {{Modelling}} of {{Bone Ageing}}}.
\newblock Doctoral, University of East Anglia
\bibAnnoteFile{davisPredictiveModellingBone2013}

\bibitem[{Deng et~al.(2009)Deng, Dong, Socher, Li, Li, and
  {Fei-Fei}}]{dengImageNetLargescaleHierarchical2009}
Deng, J., Dong, W., Socher, R., Li, L.-J., Li, K., and {Fei-Fei}, L. (2009).
\newblock {{ImageNet}}: {{A}} large-scale hierarchical image database.
\newblock In \emph{2009 {{IEEE Conference}} on {{Computer Vision}} and
  {{Pattern Recognition}}}.
\newblock \doi{10.1109/CVPR.2009.5206848}
\bibAnnoteFile{dengImageNetLargescaleHierarchical2009}

\bibitem[{Dua and Graff(2017)}]{duaUCIMachineLearning2017}
[Dataset] Dua, D. and Graff, C. (2017).
\newblock {{UCI Machine Learning Repository}}
\bibAnnoteFile{duaUCIMachineLearning2017}

\bibitem[{Dudani(1976)}]{dudaniDistanceWeightedKNearestNeighborRule1976}
Dudani, S.~A. (1976).
\newblock The {{Distance-Weighted}} k-{{Nearest-Neighbor Rule}}.
\newblock \emph{IEEE Transactions on Systems, Man, and Cybernetics} SMC-6.
\newblock \doi{10.1109/TSMC.1976.5408784}
\bibAnnoteFile{dudaniDistanceWeightedKNearestNeighborRule1976}

\bibitem[{Gao and Tan(2006)}]{gaoConvertingOutputScores2006}
Gao, J. and Tan, P.-n. (2006).
\newblock Converting {{Output Scores}} from {{Outlier Detection Algorithms}}
  into {{Probability Estimates}}.
\newblock In \emph{Sixth {{International Conference}} on {{Data Mining}}
  ({{ICDM}}'06)}.
\newblock \doi{10.1109/ICDM.2006.43}
\bibAnnoteFile{gaoConvertingOutputScores2006}

\bibitem[{Geler et~al.(2016)Geler, Kurbalija, Radovanovi{\'c}, and
  Ivanovi{\'c}}]{gelerComparisonDifferentWeighting2016}
Geler, Z., Kurbalija, V., Radovanovi{\'c}, M., and Ivanovi{\'c}, M. (2016).
\newblock Comparison of different weighting schemes for the {{kNN}} classifier
  on time-series data.
\newblock \emph{Knowl Inf Syst} 48.
\newblock \doi{10.1007/s10115-015-0881-0}
\bibAnnoteFile{gelerComparisonDifferentWeighting2016}

\bibitem[{Goldberger et~al.(2000)Goldberger, Amaral, Glass, Hausdorff, Ivanov,
  Mark et~al.}]{goldbergerPhysioBankPhysioToolkitPhysioNet2000}
Goldberger, A.~L., Amaral, L. A.~N., Glass, L., Hausdorff, J.~M., Ivanov,
  P.~C., Mark, R.~G., et~al. (2000).
\newblock {{PhysioBank}}, {{PhysioToolkit}}, and {{PhysioNet}}: {{Components}}
  of a {{New Research Resource}} for {{Complex Physiologic Signals}}.
\newblock \emph{Circulation} 101.
\newblock \doi{10.1161/01.CIR.101.23.e215}
\bibAnnoteFile{goldbergerPhysioBankPhysioToolkitPhysioNet2000}

\bibitem[{Goodge et~al.(2022)Goodge, Hooi, Ng, and
  Ng}]{goodgeLUNARUnifyingLocal2022}
Goodge, A., Hooi, B., Ng, S.~K., and Ng, W.~S. (2022).
\newblock {{LUNAR}}: {{Unifying Local Outlier Detection Methods}} via {{Graph
  Neural Networks}}.
\newblock In \emph{Proceedings of the {{First MiniCon Conference}}}
\bibAnnoteFile{goodgeLUNARUnifyingLocal2022}

\bibitem[{Hautamaki et~al.(2004)Hautamaki, Karkkainen, and
  Franti}]{hautamakiOutlierDetectionUsing2004}
Hautamaki, V., Karkkainen, I., and Franti, P. (2004).
\newblock Outlier detection using k-nearest neighbour graph.
\newblock In \emph{Proceedings of the 17th {{International Conference}} on
  {{Pattern Recognition}}, 2004. {{ICPR}} 2004.} vol.~3.
\newblock \doi{10.1109/ICPR.2004.1334558}
\bibAnnoteFile{hautamakiOutlierDetectionUsing2004}

\bibitem[{Hawkins(1980)}]{hawkinsIdentificationOutliers1980}
Hawkins, D.~M. (1980).
\newblock \emph{Identification of {{Outliers}}} ({Dordrecht}: {Springer
  Netherlands}).
\newblock \doi{10.1007/978-94-015-3994-4}
\bibAnnoteFile{hawkinsIdentificationOutliers1980}

\bibitem[{Houle(2013)}]{houleDimensionalityDiscriminabilityDensity2013}
Houle, M.~E. (2013).
\newblock Dimensionality, {{Discriminability}}, {{Density}} and {{Distance
  Distributions}}.
\newblock In \emph{2013 {{IEEE}} 13th {{International Conference}} on {{Data
  Mining Workshops}}}.
\newblock \doi{10.1109/ICDMW.2013.139}
\bibAnnoteFile{houleDimensionalityDiscriminabilityDensity2013}

\bibitem[{Janssens et~al.(2012)Janssens, Husz{\'a}r, and
  Postma}]{janssensStochasticOutlierSelection2012}
Janssens, J., Husz{\'a}r, F., and Postma, E. (2012).
\newblock \emph{Stochastic {{Outlier Selection}}}.
\newblock Tech. Rep. TiCC TR 2012\textendash 001, {Tilburg University}
\bibAnnoteFile{janssensStochasticOutlierSelection2012}

\bibitem[{Jiang et~al.(2018)Jiang, Kim, Guan, and
  Gupta}]{jiangTrustNotTrust2018}
Jiang, H., Kim, B., Guan, M., and Gupta, M. (2018).
\newblock To {{Trust Or Not To Trust A Classifier}}.
\newblock In \emph{Advances in {{Neural Information Processing Systems}}}
  ({Curran Associates, Inc.}), vol.~31
\bibAnnoteFile{jiangTrustNotTrust2018}

\bibitem[{Jin et~al.(2006)Jin, Tung, Han, and
  Wang}]{jinRankingOutliersUsing2006}
Jin, W., Tung, A. K.~H., Han, J., and Wang, W. (2006).
\newblock Ranking {{Outliers Using Symmetric Neighborhood Relationship}}.
\newblock In \emph{Advances in {{Knowledge Discovery}} and {{Data Mining}}},
  eds. W.-K. Ng, M.~Kitsuregawa, J.~Li, and K.~Chang ({Berlin, Heidelberg}:
  {Springer}), Lecture {{Notes}} in {{Computer Science}}.
\newblock \doi{10.1007/11731139_68}
\bibAnnoteFile{jinRankingOutliersUsing2006}

\bibitem[{Kauffmann et~al.(2020)Kauffmann, Ruff, Montavon, and
  M{\"u}ller}]{kauffmannCleverHansEffect2020}
[Dataset] Kauffmann, J., Ruff, L., Montavon, G., and M{\"u}ller, K.-R. (2020).
\newblock The {{Clever Hans Effect}} in {{Anomaly Detection}}
\bibAnnoteFile{kauffmannCleverHansEffect2020}

\bibitem[{Keller et~al.(2012)Keller, Muller, and
  Bohm}]{kellerHiCSHighContrast2012}
Keller, F., Muller, E., and Bohm, K. (2012).
\newblock {{HiCS}}: {{High Contrast Subspaces}} for {{Density-Based Outlier
  Ranking}}.
\newblock In \emph{2012 {{IEEE}} 28th {{International Conference}} on {{Data
  Engineering}}}.
\newblock \doi{10.1109/ICDE.2012.88}
\bibAnnoteFile{kellerHiCSHighContrast2012}

\bibitem[{Keogh et~al.(2006)Keogh, Wei, Xi, Lonardi, Shieh, and
  Sirowy}]{keoghIntelligentIconsIntegrating2006}
Keogh, E., Wei, L., Xi, X., Lonardi, S., Shieh, J., and Sirowy, S. (2006).
\newblock Intelligent {{Icons}}: {{Integrating Lite-Weight Data Mining}} and
  {{Visualization}} into {{GUI Operating Systems}}.
\newblock In \emph{Sixth {{International Conference}} on {{Data Mining}}
  ({{ICDM}}'06)}.
\newblock \doi{10.1109/ICDM.2006.90}
\bibAnnoteFile{keoghIntelligentIconsIntegrating2006}

\bibitem[{Kirner et~al.(2017)Kirner, Schubert, and
  Zimek}]{kirnerGoodBadNeighborhood2017}
Kirner, E., Schubert, E., and Zimek, A. (2017).
\newblock Good and {{Bad Neighborhood Approximations}} for {{Outlier Detection
  Ensembles}}.
\newblock \emph{Lecture Notes in Computer Science} 10609.
\newblock \doi{10.1007/978-3-319-68474-1_12}
\bibAnnoteFile{kirnerGoodBadNeighborhood2017}

\bibitem[{Knorr and Ng(1997)}]{knorrUnifiedApproachMining1997}
Knorr, E.~M. and Ng, R.~T. (1997).
\newblock A {{Unified Approach}} for {{Mining Outliers}}.
\newblock In \emph{{{CASCON}} '97: {{Proceedings}} of the 1997 {{Conference}}
  of the {{Centre}} for {{Advanced Studies}} on {{Collaborative Research}}}
  ({IBM Press}), {{CASCON}} '97.
\newblock \doi{10.5555/782010.782021}
\bibAnnoteFile{knorrUnifiedApproachMining1997}

\bibitem[{Knorr and Ng(1998)}]{knorrAlgorithmsMiningDistanceBased1998}
Knorr, E.~M. and Ng, R.~T. (1998).
\newblock Algorithms for {{Mining Distance-Based Outliers}} in {{Large
  Datasets}}.
\newblock In \emph{Proceedings of the 24rd {{International Conference}} on
  {{Very Large Data Bases}}} ({San Francisco, CA, USA}: {Morgan Kaufmann
  Publishers Inc}), {{VLDB}} '98
\bibAnnoteFile{knorrAlgorithmsMiningDistanceBased1998}

\bibitem[{Knorr et~al.(2000)Knorr, Ng, and
  Tucakov}]{knorrDistancebasedOutliersAlgorithms2000}
Knorr, E.~M., Ng, R.~T., and Tucakov, V. (2000).
\newblock Distance-based outliers: Algorithms and applications.
\newblock \emph{The VLDB Journal The International Journal on Very Large Data
  Bases} 8.
\newblock \doi{10.1007/s007780050006}
\bibAnnoteFile{knorrDistancebasedOutliersAlgorithms2000}

\bibitem[{Kriegel et~al.(2009{\natexlab{a}})Kriegel, Kr{\"o}ger, Schubert, and
  Zimek}]{kriegelLoOPLocalOutlier2009}
Kriegel, H.-P., Kr{\"o}ger, P., Schubert, E., and Zimek, A.
  (2009{\natexlab{a}}).
\newblock {{LoOP}}: {{Local Outlier Probabilities}}.
\newblock In \emph{Proceedings of the 18th {{ACM Conference}} on
  {{Information}} and {{Knowledge Management}}}, eds. D.~W.-L. Cheung, I.-Y.
  Song, W.~W. Chu, X.~Hu, and J.~J. Lin ({New York, NY, USA}: {Association for
  Computing Machinery}).
\newblock \doi{10.1145/1645953.1646195}
\bibAnnoteFile{kriegelLoOPLocalOutlier2009}

\bibitem[{Kriegel et~al.(2009{\natexlab{b}})Kriegel, Kr{\"o}ger, Schubert, and
  Zimek}]{kriegelOutlierDetectionAxisParallel2009}
Kriegel, H.-P., Kr{\"o}ger, P., Schubert, E., and Zimek, A.
  (2009{\natexlab{b}}).
\newblock Outlier {{Detection}} in {{Axis-Parallel Subspaces}} of {{High
  Dimensional Data}}.
\newblock In \emph{Advances in {{Knowledge Discovery}} and {{Data Mining}}},
  eds. T.~Theeramunkong, B.~Kijsirikul, N.~Cercone, and T.-B. Ho ({Berlin,
  Heidelberg}: {Springer Berlin Heidelberg}), Lecture {{Notes}} in {{Computer
  Science}}.
\newblock \doi{10.1007/978-3-642-01307-2_86}
\bibAnnoteFile{kriegelOutlierDetectionAxisParallel2009}

\bibitem[{Kriegel et~al.(2011)Kriegel, Kroger, Schubert, and
  Zimek}]{kriegelInterpretingUnifyingOutlier2011}
Kriegel, H.-P., Kroger, P., Schubert, E., and Zimek, A. (2011).
\newblock Interpreting and {{Unifying Outlier Scores}}.
\newblock In \emph{Proceedings of the 2011 {{SIAM International Conference}} on
  {{Data Mining}}}, eds. B.~Liu, H.~Liu, C.~Clifton, T.~Washio, and C.~Kamath
  ({Philadelphia, PA, USA}: {Society for Industrial and Applied Mathematics}).
\newblock \doi{10.1137/1.9781611972818.2}
\bibAnnoteFile{kriegelInterpretingUnifyingOutlier2011}

\bibitem[{Kriegel et~al.(2012)Kriegel, Kr{\"o}ger, Schubert, and
  Zimek}]{kriegelOutlierDetectionArbitrarily2012}
Kriegel, H.-P., Kr{\"o}ger, P., Schubert, E., and Zimek, A. (2012).
\newblock Outlier {{Detection}} in {{Arbitrarily Oriented Subspaces}}.
\newblock In \emph{2012 {{IEEE}} 12th {{International Conference}} on {{Data
  Mining}}}.
\newblock \doi{10.1109/ICDM.2012.21}
\bibAnnoteFile{kriegelOutlierDetectionArbitrarily2012}

\bibitem[{Lapuschkin et~al.(2019)Lapuschkin, W{\"a}ldchen, Binder, Montavon,
  Samek, and M{\"u}ller}]{lapuschkinUnmaskingCleverHans2019}
Lapuschkin, S., W{\"a}ldchen, S., Binder, A., Montavon, G., Samek, W., and
  M{\"u}ller, K.-R. (2019).
\newblock Unmasking {{Clever Hans}} predictors and assessing what machines
  really learn.
\newblock \emph{Nat Commun} 10.
\newblock \doi{10.1038/s41467-019-08987-4}
\bibAnnoteFile{lapuschkinUnmaskingCleverHans2019}

\bibitem[{Latecki et~al.(2007)Latecki, Lazarevic, and
  Pokrajac}]{lateckiOutlierDetectionKernel2007}
Latecki, L.~J., Lazarevic, A., and Pokrajac, D. (2007).
\newblock Outlier {{Detection}} with {{Kernel Density Functions}}.
\newblock In \emph{Machine {{Learning}} and {{Data Mining}} in {{Pattern
  Recognition}}}, ed. P.~Perner ({Berlin, Heidelberg}: {Springer}), Lecture
  {{Notes}} in {{Computer Science}}.
\newblock \doi{10.1007/978-3-540-73499-4_6}
\bibAnnoteFile{lateckiOutlierDetectionKernel2007}

\bibitem[{Lee and See(2004)}]{leeTrustAutomationDesigning2004}
Lee, J.~D. and See, K.~A. (2004).
\newblock Trust in {{Automation}}: {{Designing}} for {{Appropriate Reliance}}.
\newblock \emph{Hum Factors} 46.
\newblock \doi{10.1518/hfes.46.1.50_30392}
\bibAnnoteFile{leeTrustAutomationDesigning2004}

\bibitem[{Lellouche and
  Souris(2020)}]{lelloucheDistributionDistancesElements2020}
Lellouche, S. and Souris, M. (2020).
\newblock Distribution of {{Distances}} between {{Elements}} in a {{Compact
  Set}}.
\newblock \emph{Stats} 3.
\newblock \doi{10.3390/stats3010001}
\bibAnnoteFile{lelloucheDistributionDistancesElements2020}

\bibitem[{Li et~al.(2022)Li, Xiong, Li, Wu, Zhang, Liu
  et~al.}]{liInterpretableDeepLearning2022}
Li, X., Xiong, H., Li, X., Wu, X., Zhang, X., Liu, J., et~al. (2022).
\newblock Interpretable deep learning: Interpretation, interpretability,
  trustworthiness, and beyond.
\newblock \emph{Knowl Inf Syst} 64.
\newblock \doi{10.1007/s10115-022-01756-8}
\bibAnnoteFile{liInterpretableDeepLearning2022}

\bibitem[{Liu et~al.(2009)Liu, Zhong, Wickramasuriya, and
  Vasudevan}]{liuUWaveAccelerometerbasedPersonalized2009}
Liu, J., Zhong, L., Wickramasuriya, J., and Vasudevan, V. (2009).
\newblock {{uWave}}: {{Accelerometer-based}} personalized gesture recognition
  and its applications.
\newblock \emph{Pervasive and Mobile Computing} 5.
\newblock \doi{10.1016/j.pmcj.2009.07.007}
\bibAnnoteFile{liuUWaveAccelerometerbasedPersonalized2009}

\bibitem[{Macha and Akoglu(2018)}]{machaExplainingAnomaliesGroups2018}
Macha, M. and Akoglu, L. (2018).
\newblock Explaining anomalies in groups with characterizing subspace rules.
\newblock \emph{Data Min. Knowl. Discov.} 32.
\newblock \doi{10.1007/s10618-018-0585-7}
\bibAnnoteFile{machaExplainingAnomaliesGroups2018}

\bibitem[{Macleod et~al.(1987)Macleod, Luk, and
  Titterington}]{macleodReExaminationDistanceWeightedKNearest1987}
Macleod, J. E.~S., Luk, A., and Titterington, D.~M. (1987).
\newblock A {{Re-Examination}} of the {{Distance-Weighted}} k-{{Nearest
  Neighbor Classification Rule}}.
\newblock \emph{IEEE Transactions on Systems, Man, and Cybernetics} 17.
\newblock \doi{10.1109/TSMC.1987.289362}
\bibAnnoteFile{macleodReExaminationDistanceWeightedKNearest1987}

\bibitem[{Micenkov{\'a} et~al.(2015)Micenkov{\'a}, {\noopsort{beusekom}}{van
  Beusekom}, and Shafait}]{micenkovaStampVerificationAutomated2015}
Micenkov{\'a}, B., {\noopsort{beusekom}}{van Beusekom}, J., and Shafait, F.
  (2015).
\newblock Stamp {{Verification}} for {{Automated Document Authentication}}.
\newblock In \emph{Computational Forensics: 5th {{International Workshop}},
  {{IWCF}} 2012, {{Tsukuba}}, {{Japan}}, {{November}} 11, 2012 and 6th
  {{International Workshop}}, {{IWCF}} 2014, {{Stockholm}}, {{Sweden}},
  {{August}} 24, 2014, {{Revised}} Selected Papers / {{Utpal Garain}}, {{Faisal
  Shafait}} (Eds.)}, eds. U.~Garain and F.~Shafait ({Cham}: {Springer}), vol.
  8915 of \emph{Lecture Notes in Computer Science, 0302-9743}.
\newblock \doi{10.1007/978-3-319-20125-2_11}
\bibAnnoteFile{micenkovaStampVerificationAutomated2015}

\bibitem[{Micenkov{\'a} et~al.(2013)Micenkov{\'a}, Ng, Dang, and
  Assent}]{micenkovaExplainingOutliersSubspace2013}
Micenkov{\'a}, B., Ng, R.~T., Dang, X.-H., and Assent, I. (2013).
\newblock Explaining {{Outliers}} by {{Subspace Separability}}.
\newblock In \emph{2013 {{IEEE}} 13th {{International Conference}} on {{Data
  Mining}}}.
\newblock \doi{10.1109/ICDM.2013.132}
\bibAnnoteFile{micenkovaExplainingOutliersSubspace2013}

\bibitem[{Mueen et~al.(2011)Mueen, Keogh, and
  Young}]{mueenLogicalshapeletsExpressivePrimitive2011}
Mueen, A., Keogh, E., and Young, N. (2011).
\newblock Logical-shapelets: An expressive primitive for time series
  classification.
\newblock In \emph{Proceedings of the 17th {{ACM SIGKDD}} International
  Conference on {{Knowledge}} Discovery and Data Mining} ({San Diego California
  USA}: {ACM}).
\newblock \doi{10.1145/2020408.2020587}
\bibAnnoteFile{mueenLogicalshapeletsExpressivePrimitive2011}

\bibitem[{Muhr and Affenzeller(2022{\natexlab{a}})}]{muhrHybridCPUGPU2022}
Muhr, D. and Affenzeller, M. (2022{\natexlab{a}}).
\newblock Hybrid ({{CPU}}/{{GPU}}) {{Exact Nearest Neighbors Search}} in
  {{High-Dimensional Spaces}}.
\newblock In \emph{Artificial {{Intelligence Applications}} and
  {{Innovations}}}, eds. I.~Maglogiannis, L.~Iliadis, J.~Macintyre, and
  P.~Cortez ({Cham}: {Springer International Publishing}), {{IFIP Advances}} in
  {{Information}} and {{Communication Technology}}.
\newblock \doi{10.1007/978-3-031-08337-2_10}
\bibAnnoteFile{muhrHybridCPUGPU2022}

\bibitem[{Muhr and Affenzeller(2022{\natexlab{b}})}]{muhrLittleDataOften2022}
Muhr, D. and Affenzeller, M. (2022{\natexlab{b}}).
\newblock Little data is often enough for distance-based outlier detection.
\newblock \emph{Procedia Computer Science} 200.
\newblock \doi{10.1016/j.procs.2022.01.297}
\bibAnnoteFile{muhrLittleDataOften2022}

\bibitem[{Muhr and
  Affenzeller(2022{\natexlab{c}})}]{muhrOutlierAnomalyDetection2022}
Muhr, D. and Affenzeller, M. (2022{\natexlab{c}}).
\newblock Outlier/{{Anomaly Detection}} of~{{Univariate Time Series}}: {{A
  Dataset Collection}} and~{{Benchmark}}.
\newblock In \emph{Big {{Data Analytics}} and {{Knowledge Discovery}}}, eds.
  R.~Wrembel, J.~Gamper, G.~Kotsis, A.~M. Tjoa, and I.~Khalil ({Cham}:
  {Springer International Publishing}), Lecture {{Notes}} in {{Computer
  Science}}.
\newblock \doi{10.1007/978-3-031-12670-3_14}
\bibAnnoteFile{muhrOutlierAnomalyDetection2022}

\bibitem[{Muhr et~al.(2022)Muhr, Affenzeller, and
  Blaom}]{muhrOutlierDetectionJlModular2022}
[Dataset] Muhr, D., Affenzeller, M., and Blaom, A.~D. (2022).
\newblock {{OutlierDetection}}.jl: {{A}} modular outlier detection ecosystem
  for the {{Julia}} programming language
\bibAnnoteFile{muhrOutlierDetectionJlModular2022}

\bibitem[{Murray et~al.(2015)Murray, Liao, Stankovic, Stankovic,
  {Hauxwell-Baldwin}, Wilson et~al.}]{murrayDataManagementPlatform2015}
Murray, D., Liao, J., Stankovic, L., Stankovic, V., {Hauxwell-Baldwin}, R.,
  Wilson, C., et~al. (2015).
\newblock A data management platform for personalised real-time energy
  feedback.
\newblock In \emph{8th {{International Conference}} on {{Energy Efficiency}} in
  {{Domestic Appliances}} and {{Lighting}}} ({GBR}: {IET})
\bibAnnoteFile{murrayDataManagementPlatform2015}

\bibitem[{Olszewski(2001)}]{olszewskiGeneralizedFeatureExtraction2001}
Olszewski, R.~T. (2001).
\newblock \emph{Generalized Feature Extraction for Structural Pattern
  Recognition in Time-Series Data}.
\newblock Ph.D. thesis, Carnegie Mellon University, {USA}
\bibAnnoteFile{olszewskiGeneralizedFeatureExtraction2001}

\bibitem[{Ovadia et~al.(2019)Ovadia, Fertig, Ren, Nado, Sculley, Nowozin
  et~al.}]{ovadiaCanYouTrust2019}
Ovadia, Y., Fertig, E., Ren, J., Nado, Z., Sculley, D., Nowozin, S., et~al.
  (2019).
\newblock Can you trust your model' s uncertainty? {{Evaluating}} predictive
  uncertainty under dataset shift.
\newblock In \emph{Advances in {{Neural Information Processing Systems}}}
  ({Curran Associates, Inc.}), vol.~32
\bibAnnoteFile{ovadiaCanYouTrust2019}

\bibitem[{Pang et~al.(2015)Pang, Ting, and
  Albrecht}]{pangLeSiNNDetectingAnomalies2015}
Pang, G., Ting, K.~M., and Albrecht, D. (2015).
\newblock {{LeSiNN}}: {{Detecting Anomalies}} by {{Identifying Least Similar
  Nearest Neighbours}}.
\newblock In \emph{2015 {{IEEE International Conference}} on {{Data Mining
  Workshop}} ({{ICDMW}})}.
\newblock \doi{10.1109/ICDMW.2015.62}
\bibAnnoteFile{pangLeSiNNDetectingAnomalies2015}

\bibitem[{Pekalska and
  Duin(2000)}]{pekalskaClassifiersDissimilaritybasedPattern2000}
Pekalska, E. and Duin, R. (2000).
\newblock Classifiers for dissimilarity-based pattern recognition.
\newblock In \emph{Proceedings 15th {{International Conference}} on {{Pattern
  Recognition}}. {{ICPR-2000}}}. vol.~2.
\newblock \doi{10.1109/ICPR.2000.906008}
\bibAnnoteFile{pekalskaClassifiersDissimilaritybasedPattern2000}

\bibitem[{Perini et~al.(2021)Perini, Vercruyssen, and
  Davis}]{periniQuantifyingConfidenceAnomaly2021}
Perini, L., Vercruyssen, V., and Davis, J. (2021).
\newblock Quantifying the {{Confidence}} of {{Anomaly Detectors}} in {{Their
  Example-Wise Predictions}}.
\newblock In \emph{Machine {{Learning}} and {{Knowledge Discovery}} in
  {{Databases}}}, eds. F.~Hutter, K.~Kersting, J.~Lijffijt, and I.~Valera
  ({Cham}: {Springer International Publishing}), Lecture {{Notes}} in
  {{Computer Science}}.
\newblock \doi{10.1007/978-3-030-67664-3_14}
\bibAnnoteFile{periniQuantifyingConfidenceAnomaly2021}

\bibitem[{Radovanovi{\'c} et~al.(2015)Radovanovi{\'c}, Nanopoulos, and
  Ivanovi{\'c}}]{radovanovicReverseNearestNeighbors2015}
Radovanovi{\'c}, M., Nanopoulos, A., and Ivanovi{\'c}, M. (2015).
\newblock Reverse {{Nearest Neighbors}} in {{Unsupervised Distance-Based
  Outlier Detection}}.
\newblock \emph{IEEE Transactions on Knowledge and Data Engineering} 27.
\newblock \doi{10.1109/TKDE.2014.2365790}
\bibAnnoteFile{radovanovicReverseNearestNeighbors2015}

\bibitem[{Ramaswamy et~al.(2000)Ramaswamy, Rastogi, and
  Shim}]{ramaswamyEfficientAlgorithmsMining2000}
Ramaswamy, S., Rastogi, R., and Shim, K. (2000).
\newblock Efficient {{Algorithms}} for {{Mining Outliers}} from {{Large Data
  Sets}}.
\newblock \emph{SIGMOD Rec} 29.
\newblock \doi{10.1145/335191.335437}
\bibAnnoteFile{ramaswamyEfficientAlgorithmsMining2000}

\bibitem[{Rebbapragada et~al.(2009)Rebbapragada, Protopapas, Brodley, and
  Alcock}]{rebbapragadaFindingAnomalousPeriodic2009}
Rebbapragada, U., Protopapas, P., Brodley, C.~E., and Alcock, C. (2009).
\newblock Finding anomalous periodic time series: {{An}} application to
  catalogs of periodic variable stars.
\newblock \emph{Mach Learn} 74.
\newblock \doi{10.1007/s10994-008-5093-3}
\bibAnnoteFile{rebbapragadaFindingAnomalousPeriodic2009}

\bibitem[{Roth et~al.(2022)Roth, Pemula, Zepeda, Scholkopf, Brox, and
  Gehler}]{rothTotalRecallIndustrial2022}
Roth, K., Pemula, L., Zepeda, J., Scholkopf, B., Brox, T., and Gehler, P.
  (2022).
\newblock Towards {{Total Recall}} in {{Industrial Anomaly Detection}}.
\newblock In \emph{2022 {{IEEE}}/{{CVF Conference}} on {{Computer Vision}} and
  {{Pattern Recognition}} ({{CVPR}})} ({New Orleans, LA, USA}: {IEEE}).
\newblock \doi{10.1109/CVPR52688.2022.01392}
\bibAnnoteFile{rothTotalRecallIndustrial2022}

\bibitem[{Samek et~al.(2019)Samek, Montavon, Vedaldi, Hansen, and
  M{\"u}ller}]{samekExplainableAIInterpreting2019}
Samek, W., Montavon, G., Vedaldi, A., Hansen, L.~K., and M{\"u}ller, K.-R.
  (eds.) (2019).
\newblock \emph{Explainable {{AI}}: {{Interpreting}}, {{Explaining}} and
  {{Visualizing Deep Learning}}}, vol. 11700 of \emph{Lecture {{Notes}} in
  {{Computer Science}}} ({Cham}: {Springer International Publishing}).
\newblock \doi{10.1007/978-3-030-28954-6}
\bibAnnoteFile{samekExplainableAIInterpreting2019}

\bibitem[{Sapsanis et~al.(2013)Sapsanis, Georgoulas, Tzes, and
  Lymberopoulos}]{sapsanisImprovingEMGBased2013}
Sapsanis, C., Georgoulas, G., Tzes, A., and Lymberopoulos, D. (2013).
\newblock Improving {{EMG}} based classification of basic hand movements using
  {{EMD}}.
\newblock \emph{Annu Int Conf IEEE Eng Med Biol Soc} 2013.
\newblock \doi{10.1109/EMBC.2013.6610858}
\bibAnnoteFile{sapsanisImprovingEMGBased2013}

\bibitem[{Schnitzer et~al.(2012)Schnitzer, Flexer, Schedl, and
  Widmer}]{schnitzerLocalGlobalScaling2012}
Schnitzer, D., Flexer, A., Schedl, M., and Widmer, G. (2012).
\newblock Local and {{Global Scaling Reduce Hubs}} in {{Space}}.
\newblock \emph{Journal of Machine Learning Research} 13
\bibAnnoteFile{schnitzerLocalGlobalScaling2012}

\bibitem[{Schubert et~al.(2012)Schubert, Wojdanowski, Zimek, and
  Kriegel}]{schubertEvaluationOutlierRankings2012}
Schubert, E., Wojdanowski, R., Zimek, A., and Kriegel, H.-P. (2012).
\newblock On {{Evaluation}} of {{Outlier Rankings}} and {{Outlier Scores}}.
\newblock In \emph{Proceedings of the 2012 {{SIAM International Conference}} on
  {{Data Mining}}}, eds. J.~Ghosh, H.~Liu, I.~Davidson, C.~Domeniconi, and
  C.~Kamath ({Philadelphia, PA}: {Society for Industrial and Applied
  Mathematics}).
\newblock \doi{10.1137/1.9781611972825.90}
\bibAnnoteFile{schubertEvaluationOutlierRankings2012}

\bibitem[{Schubert et~al.(2014{\natexlab{a}})Schubert, Zimek, and
  Kriegel}]{schubertGeneralizedOutlierDetection2014}
Schubert, E., Zimek, A., and Kriegel, H.-P. (2014{\natexlab{a}}).
\newblock Generalized {{Outlier Detection}} with {{Flexible Kernel Density
  Estimates}}.
\newblock In \emph{Proceedings of the 2014 {{SIAM International Conference}} on
  {{Data Mining}}}, eds. M.~Zaki, Z.~Obradovic, P.~N. Tan, A.~Banerjee,
  C.~Kamath, and S.~Parthasarathy ({Philadelphia, PA}: {Society for Industrial
  and Applied Mathematics}).
\newblock \doi{10.1137/1.9781611973440.63}
\bibAnnoteFile{schubertGeneralizedOutlierDetection2014}

\bibitem[{Schubert et~al.(2014{\natexlab{b}})Schubert, Zimek, and
  Kriegel}]{schubertLocalOutlierDetection2014}
Schubert, E., Zimek, A., and Kriegel, H.-P. (2014{\natexlab{b}}).
\newblock Local outlier detection reconsidered: A generalized view on locality
  with applications to spatial, video, and network outlier detection.
\newblock \emph{Data Min Knowl Disc} 28.
\newblock \doi{10.1007/s10618-012-0300-z}
\bibAnnoteFile{schubertLocalOutlierDetection2014}

\bibitem[{{\noopsort{stein}}{van Stein} et~al.(2016){\noopsort{stein}}{van
  Stein}, {\noopsort{leeuwen}}{van Leeuwen}, and
  B{\"a}ck}]{vansteinLocalSubspacebasedOutlier2016}
{\noopsort{stein}}{van Stein}, B., {\noopsort{leeuwen}}{van Leeuwen}, M., and
  B{\"a}ck, T. (2016).
\newblock Local subspace-based outlier detection using global neighbourhoods.
\newblock In \emph{2016 {{IEEE International Conference}} on {{Big Data}}
  ({{Big Data}})}.
\newblock \doi{10.1109/BigData.2016.7840717}
\bibAnnoteFile{vansteinLocalSubspacebasedOutlier2016}

\bibitem[{Sugiyama and Borgwardt(2013)}]{sugiyamaRapidDistanceBasedOutlier2013}
Sugiyama, M. and Borgwardt, K. (2013).
\newblock Rapid {{Distance-Based Outlier Detection}} via {{Sampling}}.
\newblock In \emph{Advances in {{Neural Information Processing Systems}}}
  ({Curran Associates, Inc.}), vol.~26
\bibAnnoteFile{sugiyamaRapidDistanceBasedOutlier2013}

\bibitem[{Sun et~al.(2005)Sun, Papadimitriou, and
  Faloutsos}]{sunOnlineLatentVariable2005}
Sun, J., Papadimitriou, S., and Faloutsos, C. (2005).
\newblock Online {{Latent Variable Detection}} in {{Sensor Networks}}.
\newblock In \emph{Proceedings of the 21st {{International Conference}} on
  {{Data Engineering}}} ({USA}: {IEEE Computer Society}), {{ICDE}} '05.
\newblock \doi{10.1109/ICDE.2005.100}
\bibAnnoteFile{sunOnlineLatentVariable2005}

\bibitem[{Tan et~al.(2017)Tan, Webb, and
  Petitjean}]{tanIndexingClassifyingGigabytes2017}
Tan, C.~W., Webb, G.~I., and Petitjean, F. (2017).
\newblock Indexing and classifying gigabytes of time series under time warping.
\newblock In \emph{Proceedings of the 2017 {{SIAM International Conference}} on
  {{Data Mining}} ({{SDM}})} ({Society for Industrial and Applied
  Mathematics}), Proceedings.
\newblock \doi{10.1137/1.9781611974973.32}
\bibAnnoteFile{tanIndexingClassifyingGigabytes2017}

\bibitem[{Tang and He(2015)}]{tangENNExtendedNearest2015}
Tang, B. and He, H. (2015).
\newblock {{ENN}}: {{Extended Nearest Neighbor Method}} for {{Pattern
  Recognition}} [{{Research Frontier}}].
\newblock \emph{IEEE Computational Intelligence Magazine} 10.
\newblock \doi{10.1109/MCI.2015.2437512}
\bibAnnoteFile{tangENNExtendedNearest2015}

\bibitem[{Tang and He(2017)}]{tangLocalDensitybasedApproach2017}
Tang, B. and He, H. (2017).
\newblock A local density-based approach for outlier detection.
\newblock \emph{Neurocomputing} 241.
\newblock \doi{10.1016/j.neucom.2017.02.039}
\bibAnnoteFile{tangLocalDensitybasedApproach2017}

\bibitem[{Trittenbach and
  B{\"o}hm(2019)}]{trittenbachDimensionbasedSubspaceSearch2019}
Trittenbach, H. and B{\"o}hm, K. (2019).
\newblock Dimension-based subspace search for outlier detection.
\newblock \emph{Int J Data Sci Anal} 7.
\newblock \doi{10.1007/s41060-018-0137-7}
\bibAnnoteFile{trittenbachDimensionbasedSubspaceSearch2019}

\bibitem[{Vinh et~al.(2016)Vinh, Chan, Romano, Bailey, Leckie, Ramamohanarao
  et~al.}]{vinhDiscoveringOutlyingAspects2016}
Vinh, N.~X., Chan, J., Romano, S., Bailey, J., Leckie, C., Ramamohanarao, K.,
  et~al. (2016).
\newblock Discovering outlying aspects in large datasets.
\newblock \emph{Data Min Knowl Disc} 30.
\newblock \doi{10.1007/s10618-016-0453-2}
\bibAnnoteFile{vinhDiscoveringOutlyingAspects2016}

\bibitem[{Wahid and Annavarapu(2021)}]{wahidNaNODNaturalNeighbourbased2021}
Wahid, A. and Annavarapu, C. S.~R. (2021).
\newblock {{NaNOD}}: {{A}} natural neighbour-based outlier detection algorithm.
\newblock \emph{Neural Comput \& Applic} 33.
\newblock \doi{10.1007/s00521-020-05068-2}
\bibAnnoteFile{wahidNaNODNaturalNeighbourbased2021}

\bibitem[{Wang et~al.(2010)Wang, Ye, Keogh, and
  Shelton}]{wangAnnotatingHistoricalArchives2010}
Wang, X., Ye, L., Keogh, E.~J., and Shelton, C.~R. (2010).
\newblock Annotating {{Historical Archives}} of {{Images}}.
\newblock \emph{IJDLS} 1.
\newblock \doi{10.4018/jdls.2010040104}
\bibAnnoteFile{wangAnnotatingHistoricalArchives2010}

\bibitem[{Wu and Jermaine(2006)}]{wuOutlierDetectionSampling2006}
Wu, M. and Jermaine, C. (2006).
\newblock Outlier detection by sampling with accuracy guarantees.
\newblock In \emph{Proceedings of the 12th {{ACM SIGKDD}} International
  Conference on {{Knowledge}} Discovery and Data Mining} ({Philadelphia PA
  USA}: {ACM}).
\newblock \doi{10.1145/1150402.1150501}
\bibAnnoteFile{wuOutlierDetectionSampling2006}

\bibitem[{Zavrel(1997)}]{zavrelEmpiricalReExaminationWeighted1997}
Zavrel, J. (1997).
\newblock An {{Empirical Re-Examination}} of {{Weighted Voting}} for k-{{NN}}.
\newblock In \emph{{{BENELEARN-97 Proceedings}} of the 7th {{Belgian-Dutch
  Conference}} on {{Machine Learning}}}, eds. W.~Daelemans, P.~Flach, and
  A.~{\noopsort{bosch}}{van den Bosch} ({Tilburg}: {Tilburg University})
\bibAnnoteFile{zavrelEmpiricalReExaminationWeighted1997}

\bibitem[{Zhang et~al.(2009)Zhang, Hutter, and
  Jin}]{zhangNewLocalDistanceBased2009}
Zhang, K., Hutter, M., and Jin, H. (2009).
\newblock A {{New Local Distance-Based Outlier Detection Approach}} for
  {{Scattered Real-World Data}}.
\newblock In \emph{Advances in {{Knowledge Discovery}} and {{Data Mining}}:
  13th {{Pacific-Asia Conference}}, {{PAKDD}} 2009 {{Bangkok}}, {{Thailand}},
  {{April}} 27-30, 2009 {{Proceedings}}}, eds. T.~Theeramunkong, B.~Kijsirikul,
  N.~Cercone, and T.-B. Ho ({Berlin, Heidelberg}: {Springer Berlin
  Heidelberg}), vol. 5476 of \emph{Lecture {{Notes}} in {{Computer Science}}.
  0302-9743}
\bibAnnoteFile{zhangNewLocalDistanceBased2009}

\bibitem[{Zhang et~al.(2015)Zhang, Lin, and
  Karim}]{zhangAnglebasedSubspaceAnomaly2015}
Zhang, L., Lin, J., and Karim, R. (2015).
\newblock An angle-based subspace anomaly detection approach to
  high-dimensional data: {{With}} an application to industrial fault detection.
\newblock \emph{Reliability Engineering \& System Safety} 142.
\newblock \doi{10.1016/j.ress.2015.05.025}
\bibAnnoteFile{zhangAnglebasedSubspaceAnomaly2015}

\bibitem[{Zhao et~al.(2019)Zhao, Nasrullah, and Li}]{zhaoPyODPythonToolbox2019}
Zhao, Y., Nasrullah, Z., and Li, Z. (2019).
\newblock {{PyOD}}: {{A}} python toolbox for scalable outlier detection.
\newblock \emph{Journal of Machine Learning Research} 20
\bibAnnoteFile{zhaoPyODPythonToolbox2019}

\bibitem[{Zhu et~al.(2016)Zhu, Feng, and
  Huang}]{zhuNaturalNeighborSelfadaptive2016}
Zhu, Q., Feng, J., and Huang, J. (2016).
\newblock Natural neighbor: {{A}} self-adaptive neighborhood method without
  parameter {{K}}.
\newblock \emph{Pattern Recognition Letters} 80.
\newblock \doi{10.1016/j.patrec.2016.05.007}
\bibAnnoteFile{zhuNaturalNeighborSelfadaptive2016}

\bibitem[{Zimek and Filzmoser(2018)}]{zimekThereBackAgain2018}
Zimek, A. and Filzmoser, P. (2018).
\newblock There and back again: {{Outlier}} detection between statistical
  reasoning and data mining algorithms.
\newblock \emph{WIREs Data Mining and Knowledge Discovery} 8.
\newblock \doi{10.1002/widm.1280}
\bibAnnoteFile{zimekThereBackAgain2018}

\bibitem[{Zimek et~al.(2013)Zimek, Gaudet, Campello, and
  Sander}]{zimekSubsamplingEfficientEffective2013}
Zimek, A., Gaudet, M., Campello, R.~J., and Sander, J. (2013).
\newblock Subsampling for efficient and effective unsupervised outlier
  detection ensembles.
\newblock In \emph{Proceedings of the 19th {{ACM SIGKDD International
  Conference}} on {{Knowledge Discovery}} and {{Data Mining}}} ({New York, NY,
  USA}: {Association for Computing Machinery}), {{KDD}} '13.
\newblock \doi{10.1145/2487575.2487676}
\bibAnnoteFile{zimekSubsamplingEfficientEffective2013}

\end{thebibliography}

\end{document}